\documentclass[journal]{IEEEtran}

\usepackage[T1]{fontenc}
\usepackage[tbtags]{amsmath}
\usepackage{amssymb,amsthm}
\usepackage{mathtools}
\usepackage{graphicx}
\usepackage{booktabs}
\usepackage{array}
\usepackage{cite}
\usepackage[hidelinks]{hyperref}

\graphicspath{{figs/}}

\newtheorem{theorem}{Theorem}
\newtheorem{proposition}{Proposition}

\newcommand{\Prob}{\mathbb{P}}
\newcommand{\E}{\mathbb{E}}
\newcommand{\R}{\mathbb{R}}
\newcommand{\TV}{\mathrm{TV}}
\newcommand{\osc}{\operatorname{osc}}
\newcommand{\Mult}{\mathrm{Mult}}
\newcommand{\Bin}{\mathrm{Bin}}
\newcommand{\Var}{\operatorname{Var}}

\newcommand{\ind}{\mathbf{1}}
\newcommand{\eqd}{\overset{d}{=}}
\newcommand{\Phibar}{\bar{\Phi}}
\newcommand{\gs}{g^{\sigma}}

\begin{document}

\title{Predicting Rare-Mode Collapse in Recursive Generative Training from Single-Generation Calibration}

\author{Jingyao~Zhang\textsuperscript{*,\dag}, Mingyu~Du\textsuperscript{*}, and Luhan%
\thanks{\textsuperscript{*}Jingyao Zhang and Mingyu Du contributed equally to this work. \textsuperscript{\dag}Corresponding author: Jingyao Zhang.}%
\thanks{Jingyao Zhang and Luhan are with the University of Sydney, Sydney, NSW, Australia (e-mail: jzha0544@sydney.edu.au).}%
\thanks{Mingyu Du is with the University of New South Wales, Sydney, NSW, Australia (e-mail: dumingyu2004@gmail.com).}%
\thanks{Code, frozen forecasts, and experiment records supporting this study will be released upon publication.}}

\markboth{IEEE TRANSACTIONS ON NEURAL NETWORKS AND LEARNING SYSTEMS (WORKING DRAFT)}%
{Predicting Rare-Mode Collapse in Recursive Generative Training}

\maketitle

\begin{abstract}
Recursive generative training can preserve sample counts in a rare region while silently eroding its characteristic core. Mapping this finite-horizon collapse across real-data quotas, rare prevalences, and feedback perturbations via exhaustive multi-generation chains is computationally prohibitive. We develop a decoupled framework that forecasts multi-generation collapse probabilities and locates operating risk boundaries from independent single-generation conditional fits alone. Pairing an exact three-source multinomial count law with single-generation response banks propagates count marginals in closed form, backed by explicit calibration-error and boundary-resolution bounds. For additive Gaussian perturbations, we further prove that equal-variance all-row and sparse policies reverse their boundary-crossing order at a unique hyperplane distance. Across a 2,178-setting count benchmark, exact propagation cuts risk mean absolute error (MAE) from 1.85 to 0.49 percentage points (pp). Across ten two-dimensional learners, calibration achieves 2.60--5.69~pp six-round core MAE on nine stationary learners (65--81\% below an unperturbed baseline), diagnoses non-stationary outlier drift in standard DDIM (16.38~pp) via a stability probe, and resolves the predicted six-round geometric reversal. On Fashion-MNIST latent flow-matching chains, calibration outperforms direct-chain Gaussian-process interpolation at lower fit budgets and tracks the 20\% latent-perturbation boundary shift (3.14 versus 10.78~pp MAE for proportion copying) driven by opposing inflow to the rare region and outflow from its core.
\end{abstract}

\begin{IEEEkeywords}
Recursive generative training, model collapse, rare-mode coverage, Gaussian perturbation, finite-horizon risk forecasting, generative models.
\end{IEEEkeywords}

\section{Introduction}\label{sec:intro}

\IEEEPARstart{A}{s} synthetic outputs accumulate across web-scale corpora, newly trained generative models increasingly ingest data produced by earlier generations~\cite{ref1,ref8}. Under recursive retraining, rare subpopulations face an asymmetric hazard: aggregate metrics over a rare category can appear stable, or even grow, while the characteristic core of the rare mode steadily disappears. For example, in our Fashion-MNIST study a classifier still labels more than a fifth of generated images as bags after six rounds, while the most typical bags have fallen to 57\% of their reference share. When feedback perturbations push majority samples across a loose class boundary while simultaneously dispersing rare samples away from their density peak, monitoring coarse region counts masks genuine mode loss. Distinguishing coarse region occupancy from core retention is therefore essential for evaluating recursive training safety.

Forecasting this collapse across deployment choices poses a severe computational bottleneck. Consider retraining an unconditional generator over six recursive generations on a 128-example pool that mixes fresh real examples with synthetic outputs from the previous model, where a rare mode starts at 5\% prevalence and synthetic feedback undergoes perturbation before retraining. System designers must navigate three interacting controls---the synthetic share, the reference rare prevalence, and the feedback perturbation scale---to keep finite-horizon core-failure risk below an allowable operating budget (formally defined in Section~\ref{sec:method}). Mapping the resulting operating boundary by executing hundreds of independent multi-generation chains at every setting requires tens of thousands of model fits, which rapidly becomes intractable for neural image generators.

\begin{figure*}[!t]
\centering
\includegraphics[width=\textwidth]{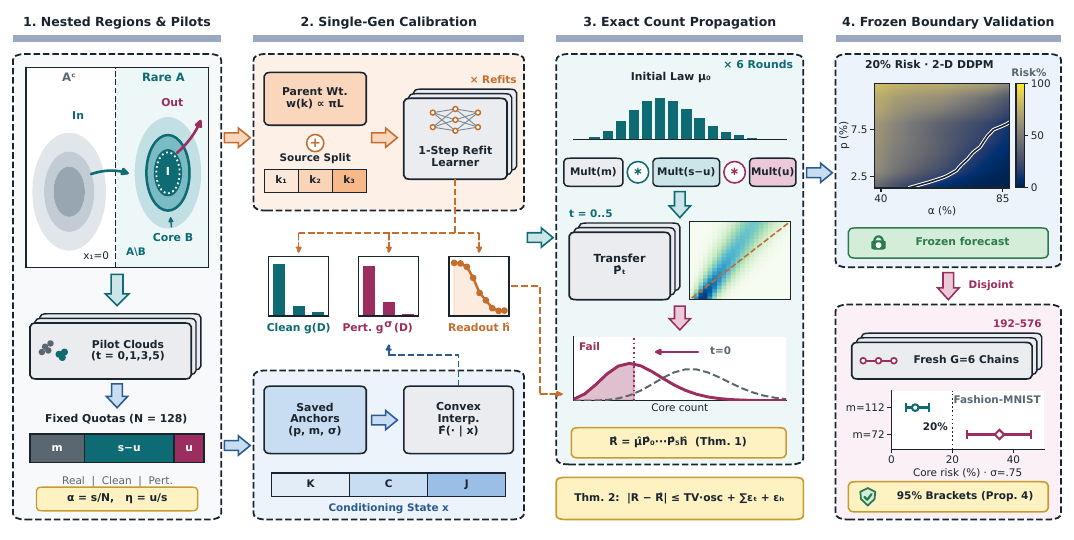}
\caption{Overview of the decoupled single-generation forecasting and validation workflow. \textbf{1.~Nested Regions \& Pilots:} Perturbation drives net inflow from common region $A^c$ across $x_1=0$ into periphery $A \setminus B$, while driving net outflow from core $B$ (solid ring; inner ellipse $I$ dotted). Short pilot chains record outputs at rounds $0,1,3,5$. The quota strip ($m=40$, quarter-row setting, $40:66:22$) shows fresh real (grey, $m$), clean synthetic (teal, $s-u$), and perturbed synthetic (plum, $u$) rows. \textbf{2.~Single-Gen Calibration:} Saved anchor runs are reweighted and split by source counts $(k_1,k_2,k_3)$ to train independent one-step models. Three mini-plots summarize the resulting outputs: teal clean $g(D)$ and plum perturbed $g^\sigma(D)$ shares across $(A^c, A \setminus B, B)$ (Fashion-MNIST, $m=72$, $\sigma=0.75$), and orange binned failure rates $\widehat{h}$. Convex interpolation across anchors supplies responses for any count state $(K,C,J)$. \textbf{3.~Exact Count Propagation:} Starting from initial count distribution $\mu_0$, convolving the three source counts produces transfer matrix $\widehat{P}_t$ (heatmap: darker teal indicates higher probability). Over six rounds, core counts shift from $t=0$ (dashed grey) leftward to $t=6$ (solid plum, shaded failure tail), giving terminal risk $\widehat{R}$ (Theorems~\ref{thm:1}--\ref{thm:2}). \textbf{4.~Frozen Boundary Validation:} The predicted 20\% risk map (2-D DDPM, Fig.~\ref{fig:4}c; colorbar shows risk \%) is locked before running 192--576 independent chains. Bottom bars test Fashion-MNIST ($p=20\%$, $\sigma=0.75$) at $m \in \{72, 112\}$, whose joint 95\% intervals bracket the dotted 20\% line (Proposition~\ref{prop:4}).}
\label{fig:1}
\end{figure*}

Existing theory on model collapse does not resolve this finite-horizon forecasting problem. Prior studies have characterized asymptotic tail absorption~\cite{ref1,ref2}, local stability near population optima~\cite{ref6}, admissible synthetic fractions under asymptotic effective sample size~\cite{ref8}, and upper bounds on accumulated score divergence in diffusion chains~\cite{ref13,ref14}. While these analyses establish that mixture ratios, sampling bias, and data retention govern long-run degradation, they do not provide calibrated, finite-horizon failure probabilities for a specified neural learner under concrete regional geometry, nor do they quantify how much single-generation calibration reduces uncertainty about an operating boundary.

In this paper, we ask whether independent single-generation conditional fits can accurately forecast multi-generation core collapse and locate operating risk boundaries across diverse learners and image domains. Figure~\ref{fig:1} illustrates our decoupled framework. Rather than unrolling full recursive chains across the entire parameter space, we separate the known finite-sample count dynamics from the unknown learner response. First, conditional on a fitted parent model, the regional counts of fresh real, clean synthetic, and perturbed synthetic rows follow an exact three-source multinomial convolution. Second, independent single-generation refits organized by training count state estimate the paired clean and perturbed output laws of the target learner. Composing these two ingredients yields a closed-form Markov forecast of terminal failure risk and its operating boundary, which we freeze before launching any independent recursive validation chains.

To probe the geometric mechanism behind hidden core erosion, we also analyze additive Gaussian feedback perturbations through paired inflow--outflow accounting and continuous heat flux (Fig.~\ref{fig:2}). Comparing two equal-variance policies---perturbing every synthetic row at scale $\sigma$ versus perturbing a fraction $a^{-2}$ of rows at scale $a\sigma$---reveals a sharp geometric test: their pointwise boundary-crossing order reverses at a unique threshold distance $d_*$, linking micro-level spatial transport directly to multi-generation risk shifts.

Our main contributions are threefold:
\begin{enumerate}
\item \textbf{Decoupled single-generation forecasting with exact error bounds.} We formulate finite-horizon regional collapse risk under fixed source quotas, prove an exact marginal-propagation representation (Theorem~\ref{thm:1}) built on the three-source conditional count law (Proposition~\ref{prop:2}), and derive an explicit propagation bound for calibration error (Theorem~\ref{thm:2}) together with a local boundary-resolution guarantee (Proposition~\ref{prop:4}).
\item \textbf{Geometric transport theory and equal-variance reversal.} We establish a continuous Gaussian heat-flux identity for regional mass change and prove a unique pointwise crossing-order reversal between equal-variance all-row and sparse perturbation policies (Proposition~\ref{prop:1}). Controlled near/far interventions verify this sign flip in one step across all ten two-dimensional learners and resolve the full six-round risk reversal in non-saturating learners.
\item \textbf{Multi-learner, failure-mode, and image-domain validation.} Under a pre-validation freeze protocol, we validate the framework across a 2,178-setting exact count benchmark, ten two-dimensional learners, and Fashion-MNIST latent flow-matching chains. Across nine stationary two-dimensional learners, calibration holds error within 2.60--5.69~pp MAE, while a multi-round stability probe diagnoses non-stationary outlier drift in standard DDIM and is evaluated prospectively on five held-out learners. On Fashion-MNIST, single-generation calibration surpasses direct-chain Gaussian-process interpolation at lower compute budgets, generalizes across failure thresholds without retraining, and accurately tracks latent-perturbation boundary shifts driven by opposing inflow to rare region $A$ and outflow from core $B$.
\end{enumerate}

The remainder of this paper is organized as follows. Section~\ref{sec:related} situates our framework within the model-collapse literature. Section~\ref{sec:method} develops the recursive protocol, regional transport theory, and decoupled forecasting guarantees. Section~\ref{sec:count_ddpm} validates the exact count law and primary two-dimensional diffusion boundaries. Section~\ref{sec:multilearner} evaluates generalization across ten learners, non-stationary drift limits, and geometric reversals. Section~\ref{sec:fmnist} presents the Fashion-MNIST image studies, compute-efficiency comparisons, and round-by-round transport measurements. Section~\ref{sec:discussion} synthesizes mechanistic insights and claim boundaries, and Section~\ref{sec:conclusion} concludes.

\section{Related Work}\label{sec:related}

\subsection{Model Collapse, Tail Erosion, and Data Retention}

Early analyses of recursive self-training established that iteratively fitting models to their own outputs erodes distribution tails and drives discrete systems toward absorbing states~\cite{ref1,ref2}. Follow-up work demonstrated that feedback loops disproportionately degrade minority group representation~\cite{ref3} and altered scaling behavior when tail truncation or temperature shifts bias synthetic sampling~\cite{ref15}. In generative evaluation, precision and recall metrics already separate sample fidelity from distribution coverage~\cite{ref4,ref5}. Our formulation builds on that distinction but targets a stricter operational question: rather than measuring zero-support absorption or aggregate fidelity, we forecast the finite-horizon probability that a generator's clean coverage of a nested core $B \subset A$ drops below a prescribed fraction of its reference mass, while allowing individual chains to recover between rounds.

A parallel theoretical line identifies conditions under which fresh real data stabilizes recursive training. Bertrand \emph{et al.}~\cite{ref6} derived local stability criteria near population optima with finite-sample extensions, Seddik \emph{et al.}~\cite{ref7} linked admissible synthetic fractions to initial distributional deviation in language models, and Alemohammad \emph{et al.}~\cite{ref8} mapped stability regions under biased self-consuming loops using asymptotic effective sample size. Several studies further showed that accumulating historical real and synthetic pools prevents complete collapse across iterative Gaussian estimation and diffusion training~\cite{ref9,ref10,ref11,ref12}. In contrast to accumulating-pool analyses and asymptotic stability thresholds, our fixed-budget protocol trains each successor on fixed quotas of fresh real and immediate-parent outputs and converts single-generation measurements into calibrated, finite-horizon failure probabilities on an explicit $(\alpha,\sigma,p)$ grid.

\subsection{Feedback Perturbations, Diffusion Dynamics, and Sampling Bias}

Recent analyses of score-based diffusion under recursion bound accumulated distributional divergence in terms of fresh-data fractions and score approximation error~\cite{ref13}, and isolate residual smoothing induced by early stopping of reverse samplers~\cite{ref14}. Other works examine how entropy-based selection mediates the transition from generalization to memorization~\cite{ref16} or how filtering against local references preferentially eliminates global tail modes~\cite{ref17}.

Our approach complements these studies in two ways. First, we introduce a value-independent feedback perturbation at the interface between generations, allowing the spatial transport kernel to be varied independently of the reverse sampler and real quota. Second, by pairing two equal-variance Gaussian policies that differ only in tail weight, we isolate pure geometric transport from perturbation variance and connect one-step regional flux directly to multi-generation core-failure risk across ten two-dimensional learners and image latent spaces.

\section{Protocol, Regional Transport, and Decoupled Forecasting}\label{sec:method}

This section develops our mathematical framework in four linked modules: (i) the fixed-quota recursive protocol and finite-horizon risk objective, (ii) paired regional transport and equal-variance crossing reversal, (iii) the exact three-source conditional count law, and (iv) the decoupled single-generation forecasting engine with calibration-error and boundary-resolution guarantees.

\subsection{Recursive Protocol and Finite-Horizon Collapse Risk}

Let $\mathcal{X}$ denote the sample space and let $H_-$ and $H_+$ be the conditional distributions of the majority (common) and rare subpopulations. The reference data distribution is the mixture $H_p = (1-p)H_- + pH_+$, indexed by the reference rare prevalence $p \in (0,1)$. We fix a measurable \emph{rare region} $A \subseteq \mathcal{X}$ and a strictly smaller \emph{rare core} $B \subset A$ capturing the most characteristic examples of the rare mode. Together, these regions partition $\mathcal{X}$ into $D_1 = A^c$, $D_2 = A \setminus B$, and $D_3 = B$. In our two-dimensional benchmark, $H_-$ and $H_+$ are supported on $A^c = \{x_1 \le 0\}$ and $A = \{x_1 > 0\}$, respectively, with $B$ an ellipse centered on the rare peak, yielding $H_p(A)=p$ and $H_p(B)=pH_+(B)$. In our image study, a fixed classifier defines both $A$ and $B$, and reference coverage follows the full mixture identity $H_p(D) = (1-p)H_-(D) + pH_+(D)$ for $D \in \{A,B\}$.

\begin{table}[!t]
\caption{Canonical Notation for the Recursive Protocol, Transport Law, and Risk Boundary.}
\label{tab:1}
\centering
\begin{tabular}{@{}lp{0.68\columnwidth}@{}}
\toprule
Symbol & Definition \\
\midrule
$N,m,s,u$ & Total, fresh real, synthetic, and perturbed-synthetic rows per round \\
$\alpha,\eta$ & Synthetic share $s/N = 1-m/N$ and perturbed synthetic share $u/s$ \\
$p,\sigma$ & Reference rare prevalence and feedback perturbation strength \\
$A,B,I$ & Rare region, nested core ($B \subset A$), and inner ellipse ($I \subset B$) \\
$K_t,C_t,J_t$ & Round-$t$ training counts in $A$, $B$, and $I$; state vector $X_t$ \\
$M_t(D)$ & Clean population coverage of region $D$ under fitted generator $M_t$ \\
$g_t,g_t^\sigma$ & Clean and perturbed regional probability vectors of parent $M_t$ \\
$G,\gamma,\beta$ & Retraining horizon, retained coverage fraction, and risk budget \\
$R_G^D,\mathcal{B}^D$ & Terminal failure risk at round $G$ and operating boundary $\{R_G^D=\beta\}$ \\
\bottomrule
\end{tabular}
\end{table}

At round $t=0$, an initial generator $M_0$ is fitted to $N$ independent real draws from $H_p$. For each subsequent generation $t \ge 0$, the training dataset for successor $M_{t+1}$ is assembled from fixed source quotas: $m$ fresh real samples drawn independently from $H_p$ and $s = N - m$ synthetic samples drawn independently from the fitted parent $M_t$. Among the $s$ synthetic rows, exactly $u$ rows are selected uniformly at random (independent of their feature values) and passed through a perturbation kernel of strength $\sigma$, where $\sigma=0$ denotes no perturbation. The remaining $s-u$ synthetic rows and all $m$ fresh rows enter unchanged. Define the synthetic share $\alpha = s/N$ and perturbed synthetic share $\eta = u/s$ (with $\eta=0$ when $s=0$). All learners are trained unconditionally on feature vectors without rare-class labels.

For the training set that produces $M_t$, let $K_t$ and $C_t$ count rows falling in $A$ and $B$, respectively, forming the partition count vector $Z_t = (N-K_t, K_t-C_t, C_t)$. Let $X_t$ denote the chosen discrete count state (such as $K_t$, $(K_t,C_t)$, or $(K_t,C_t,J_t)$ when an inner ellipse $I \subset B$ is tracked). Once fitted, $M_t$ is a random probability measure, and $M_t(D)$ denotes its clean population probability of generating a sample in region $D$.

For $D \in \{A,B\}$ with $H_p(D)>0$, fix a retraining horizon $G$ and retained coverage fraction $\gamma \in (0,1)$, setting the primary failure cutoff $\tau_D(p) = \gamma H_p(D)$. A chain fails at horizon $G$ if its clean terminal coverage falls below $\tau_D(p)$:
\begin{equation}\label{eq:1}
Y_G^D = \ind\{M_G(D) < \tau_D(p)\}, \qquad R_G^D = \E[Y_G^D],
\end{equation}
where the expectation integrates over initial data, feedback draws, perturbations, and training randomness. Because $Y_G^D$ evaluates clean coverage strictly at round $G$, chains that dip below $\tau_D(p)$ earlier may recover. Given an allowable risk budget $\beta \in (0,1)$, the \emph{operating risk boundary} is the level set
\begin{equation}\label{eq:boundary}
\mathcal{B}^D = \{(\alpha,\sigma,p) : R_G^D(\alpha,\sigma,p) = \beta\}.
\end{equation}
Throughout our primary experiments, $N=128$, $G=6$, $\gamma=1/2$, and $\beta=0.20$. Table~\ref{tab:1} summarizes the locked notation.

\subsection{Regional Transport and Equal-Variance Crossing Reversal}

To explain why a single feedback intervention can enlarge rare region $A$ while depleting core $B$, hold a fitted parent $M$ fixed and couple a clean draw $V \sim M$ with its perturbed counterpart $\widetilde{V}$. Writing $g(D) = \Prob(V \in D \mid M)$ and $\gs(D) = \Prob(\widetilde{V} \in D \mid M)$, subtracting the two probabilities cancels rows that remain inside $D$:
\begin{equation}\label{eq:2}
\begin{split}
\gs(D) - g(D) ={} & \Prob(V \notin D, \widetilde{V} \in D \mid M) \\
& - \Prob(V \in D, \widetilde{V} \notin D \mid M).
\end{split}
\end{equation}
Equation~\eqref{eq:2} decomposes net regional mass change into \emph{inflow} minus \emph{outflow}. This exact identity applies both to ambient-space Gaussian displacements $\widetilde{V} = V + \sigma\xi$ ($\xi \sim \mathcal{N}(0,I_d)$) and to latent-space image perturbations applied prior to decoding.

\begin{figure*}[!t]
\centering
\includegraphics[width=0.85\textwidth]{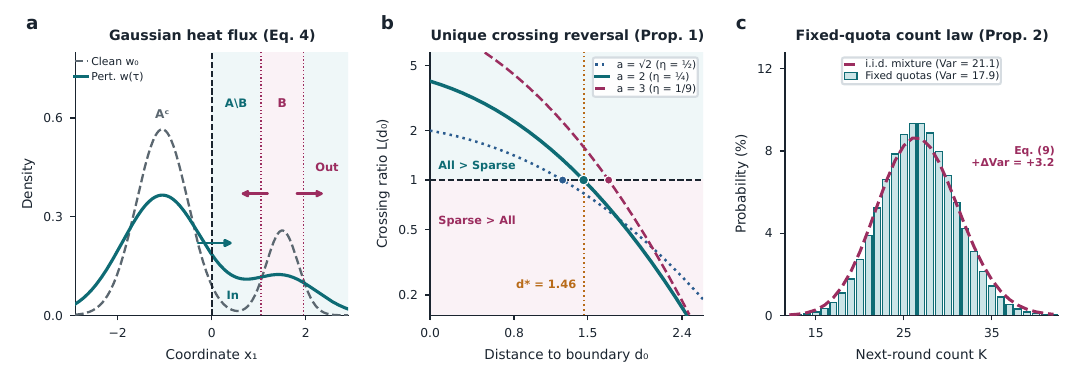}
\caption{Analytical mechanisms underlying regional transport and finite-sample count dynamics in Section~\ref{sec:method}. \textbf{(a)~Gaussian heat flux (Eq.~\eqref{eq:3}):} smoothing a bimodal clean density $w_0$ (dashed slate) to $w(\tau)$ (solid teal) pushes majority mass across $x_1=0$ into periphery $A \setminus B$ (net inflow) while flattening the narrow rare peak $B$ outward (net outflow). \textbf{(b)~Unique crossing-order reversal (Proposition~\ref{prop:1}, Eq.~\eqref{eq:4}):} the all-row to sparse crossing ratio $L(d_0) = a^2\Phibar(d_0/\sigma)/\Phibar(d_0/(a\sigma))$ ($\sigma=0.90$) decreases monotonically across $L=1$ at a unique threshold $d_*(a)$ for each sparsity multiplier $a \in \{\sqrt{2}, 2, 3\}$ ($d_* = 1.46$ at $a=2$), flipping dominance from all-row ($L>1$) to sparse ($L<1$). \textbf{(c)~Fixed-quota count law versus i.i.d.\ mixture (Proposition~\ref{prop:2}, Eq.~\eqref{eq:8}):} exact three-source multinomial convolution (teal bars, $\Var = 17.9$) produces tighter count tails than a mean-matched i.i.d.\ synthetic mixture (dashed plum, $\Var = 21.1$, inflating variance by $s\eta(1-\eta)(g^\sigma-g)^2 = +3.2$ at $N=128$, $m=16$, $\eta=1/2$, $g=0.38$, $g^\sigma=0.04$).}
\label{fig:2}
\end{figure*}

In Euclidean space $\R^d$, setting $\tau = \sigma^2$ makes the perturbed density $w_\tau = M * \varphi_\tau$ evolve according to the heat equation $\partial_\tau w_\tau = \frac{1}{2}\Delta w_\tau$. For any bounded region $D$ with $C^1$ boundary $\partial D$ and outward unit normal $n$, differentiating under the integral sign for $\tau > 0$ and applying the divergence theorem gives the continuous regional flux identity:
\begin{equation}\label{eq:3}
\begin{split}
\frac{d}{d\tau} g^{\sqrt{\tau}}(D) &= \frac{1}{2} \int_D \Delta w_\tau(x)\,dx \\
&= \frac{1}{2} \int_{\partial D} \nabla w_\tau(x)\cdot n(x)\,dS(x).
\end{split}
\end{equation}
Around a localized rare peak $B$, density decreases outward ($\nabla w_\tau \cdot n < 0$), ensuring net outflow from $B$. Simultaneously, across a broader decision boundary $\partial A$ adjacent to a high-density majority mode, the density gradient can point inward, producing positive net inflow into $A$ (Fig.~\ref{fig:2}a).

This geometric sensitivity yields a decisive experimental test. For a randomly chosen synthetic row displaced by $\sigma\xi$ with probability $\eta$, the per-coordinate perturbation variance is $\eta\sigma^2$. Let $\phi$, $\Phi$, and $\Phibar = 1-\Phi$ denote the standard normal density, cumulative distribution, and upper tail. A point at perpendicular distance $d_0 > 0$ from a separating hyperplane crosses under displacement scale $\sigma$ with probability $\Phibar(d_0/\sigma)$.

\begin{proposition}[Unique reversal of crossing order]\label{prop:1}
Fix $\sigma > 0$ and $a > 1$. Compare perturbing every synthetic row at scale $\sigma$ ($\eta = 1$) against perturbing an attainable fraction $\eta = a^{-2}$ of rows at scale $a\sigma$, matching per-row variance $\eta\sigma^2$. Over a uniformly selected synthetic row at distance $d_0 > 0$ from a hyperplane, there exists a unique threshold $d_* > 0$ at which the two policies have equal crossing probability. The all-row policy crosses more often for $0 < d_0 < d_*$, and the sparse policy crosses more often for $d_0 > d_*$.
\end{proposition}

\begin{proof}
Let $z = d_0/\sigma > 0$. Because unperturbed rows cannot cross, the ratio of all-row to sparse crossing probabilities is
\begin{equation}\label{eq:4}
L(z) = \frac{a^2 \Phibar(z)}{\Phibar(z/a)}.
\end{equation}
The standard normal hazard rate $\lambda_{\mathrm{N}}(z) = \phi(z)/\Phibar(z)$ satisfies $\Phibar(z) < \phi(z)/z$ for $z>0$, hence $\lambda_{\mathrm{N}}'(z) = \lambda_{\mathrm{N}}(z)[\lambda_{\mathrm{N}}(z)-z] > 0$. Therefore $\frac{d}{dz}\log L(z) = -\lambda_{\mathrm{N}}(z) + a^{-1}\lambda_{\mathrm{N}}(z/a) < 0$, so $L$ is strictly decreasing on $(0,\infty)$. Substituting $t = av$ in $\Phibar(z)$ yields $L(z) \le a^3\exp[-(a^2-1)z^2/(2a^2)] \to 0$ as $z \to \infty$. Since $L(0)=a^2>1$ and $L$ is continuous, $L(z)=1$ has a unique positive root $z_*$, giving $d_* = \sigma z_*$.
\end{proof}

As illustrated in Fig.~\ref{fig:2}b, Proposition~\ref{prop:1} predicts that matching perturbation variance does not determine transport: shifting component centers across $d_*$ reverses which equal-variance policy generates greater cross-boundary flux. For symmetric conditional distance distributions with common perturbed crossing probability $\ell$, a parent with $q = g(A)$ contributes expected synthetic rare mass $q + \eta\ell(1-2q)$, showing how pointwise reversal propagates into net regional composition.

\subsection{Exact Three-Source Count Law and Row Substitution}

To avoid Gaussian or mean-field approximations that distort rare-event tails, we characterize the exact distribution of next-generation training counts. Let $r = (H_p(D_j))_j$, $g_t = (M_t(D_j))_j$, and $g_t^\sigma$ denote the regional probability vectors of fresh real, clean synthetic, and perturbed synthetic rows, respectively. Write $\Mult(n,b)$ for a multinomial count vector with $n$ trials and probabilities $b$, and $\Bin(n,q)$ for a scalar binomial count.

\begin{proposition}[Exact three-source conditional count law]\label{prop:2}
Under the fixed quotas and conditional row independence of Section~\ref{sec:method}-A,
\begin{equation}\label{eq:5}
Z_{t+1} \mid M_t \;\eqd\; \Mult(m,r) + \Mult(s-u, g_t) + \Mult(u, g_t^\sigma),
\end{equation}
where the three multinomial vectors on the right are mutually independent. Equivalently, the probability mass at $z$ is the coefficient of $\prod_j v_j^{z_j}$ in the probability generating function
\begin{equation}\label{eq:6}
\E\Biggl[\prod_j v_j^{Z_{t+1,j}} \;\Big|\; M_t\Biggr] = (r^\top v)^m (g_t^\top v)^{s-u} ((g_t^\sigma)^\top v)^u.
\end{equation}
\end{proposition}

\begin{proof}
Condition on both $M_t$ and the subset of $u$ perturbed row indices. Conditional independence across rows produces three independent multinomial sums whose distributions depend on the selected indices only through the cardinality $u$ by exchangeability. Averaging over the uniform index selection preserves the exact law~\eqref{eq:5}, and multiplying the single-row generating functions $b^\top v$ across independent rows yields~\eqref{eq:6}.
\end{proof}

Taking expectations in~\eqref{eq:5} separates three drivers of training-set drift:
\begin{equation}\label{eq:7}
\E[Z_{t+1}/N \mid M_t] - r = \alpha\bigl[(g_t - r) + \eta(g_t^\sigma - g_t)\bigr],
\end{equation}
namely the synthetic share $\alpha$, inherited parent bias $(g_t-r)$, and perturbation transport $\eta(g_t^\sigma-g_t)$. Crucially, replacing the fixed-quota convolution~\eqref{eq:5} with an i.i.d.\ mixture $\Mult(s, (1-\eta)g_t + \eta g_t^\sigma)$ preserves the mean~\eqref{eq:7} but strictly inflates the covariance by the positive semidefinite matrix
\begin{equation}\label{eq:8}
s\eta(1-\eta)(g_t^\sigma - g_t)(g_t^\sigma - g_t)^\top,
\end{equation}
confirming that preserving fixed quotas matters for finite-sample risk tails (Fig.~\ref{fig:2}c).

When evaluating incremental interventions, the same conditional structure isolates how perturbing one additional synthetic row shifts any count-dependent functional $v$.

\begin{proposition}[One-row substitution identity]\label{prop:3}
Holding $M_t$ and all other quotas fixed, replace one clean synthetic row by a perturbed row and let $S$ be the count vector from the remaining $N-1$ rows. For any real-valued function $v$, the change in $\E[v(Z_{t+1}) \mid M_t]$ equals
\begin{equation}\label{eq:9}
\sum_j (g_{t,j}^\sigma - g_{t,j})\,\E[v(S + e_j) \mid M_t].
\end{equation}
For the scalar count $K$ with $q_t = g_t(A)$ and $q_t^\sigma = g_t^\sigma(A)$, this simplifies to
\begin{equation}\label{eq:10}
(q_t^\sigma - q_t)\,\E[v(S+1) - v(S) \mid M_t].
\end{equation}
\end{proposition}

\begin{proof}
Condition on $(M_t, S)$, expand the expectation over the regional assignment of the substituted row under $g_t^\sigma$ and $g_t$, and take the difference.
\end{proof}

\subsection{Decoupled Risk Propagation, Calibration, and Guarantees}

\emph{Motivation.} Direct Monte Carlo estimation of $R_G^D(\alpha,\sigma,p)$ couples data sampling with repeated neural retraining at every $(\alpha,\sigma,p)$ point. Our forecasting engine decouples these two components: we propagate the finite count state $X_t \in \mathcal{S}$ analytically while estimating the learner's one-step conditional response from independent single-generation refits.

\emph{Exact propagation.} Let $\mu_t(x) = \Prob(X_t = x)$ (row vector, with $\mu = \mu_0$) and write $U_t = (g_t, g_t^\sigma)$ for the parent's paired clean and perturbed regional probabilities. Let $F_t(dv \mid x)$ be the true conditional law of $U_t$ given $X_t = x$, and let $B_t(v,y)$ be the probability of next state $y$ obtained by projecting~\eqref{eq:5} onto $X_{t+1}$. Define the state transition matrix $P_t$ and terminal readout column vector $h_G^D$ by
\begin{align}
P_t(x,y) &= \int B_t(v,y)\,F_t(dv \mid x), \label{eq:11}\\
h_G^D(x) &= \Prob\bigl(M_G(D) < \tau_D(p) \mid X_G = x\bigr). \label{eq:12}
\end{align}

\begin{theorem}[Exact terminal-risk representation]\label{thm:1}
Under Proposition~\ref{prop:2}, the finite-horizon collapse risk~\eqref{eq:1} satisfies
\begin{equation}\label{eq:13}
R_G^D = \mu P_0 P_1 \cdots P_{G-1} h_G^D
\end{equation}
for every choice of finite count state $X_t \in \mathcal{S}$.
\end{theorem}

\begin{proof}
By iterated expectation, $\Prob(X_{t+1}=y \mid X_t=x) = \int B_t(v,y)\,F_t(dv\mid x) = P_t(x,y)$, so $\mu_{t+1} = \mu_t P_t$ and inductively $\mu_G = \mu P_0 \cdots P_{G-1}$. Conditioning the terminal failure indicator on $X_G$ yields $R_G^D = \sum_x \mu_G(x) h_G^D(x) = \mu_G h_G^D$.
\end{proof}

Theorem~\ref{thm:1} does not assume that the recursive chain is Markov in $X_t$; rather, it is an exact identity for one-time marginals when $P_t$ uses the true depth-specific conditional response law $F_t$. Approximating $F_t$ across depths by a pooled single-generation response bank is an explicit modeling step whose error we bound below.

\emph{Single-generation conditional calibration.} To build the empirical response bank $\widehat{F}$ without running validation chains, we first run a small set of disjoint pilot chains at sparse calibration anchors and save parent output clouds $\{j\}$ with prior weights $\pi_j$. Given a target training count $k$ in region $A$, let $L_j(k)$ be the probability of total count $k$ under parent $j$ from~\eqref{eq:5} with source sizes $(n_1,n_2,n_3) = (m, s-u, u)$ and rare-region probabilities $(q_{j,1}, q_{j,2}, q_{j,3}) = (H_p(A), g_j(A), g_j^\sigma(A))$. Bayes' rule assigns conditional parent weights
\begin{equation}\label{eq:14}
w_j(k) = \frac{\pi_j L_j(k)}{\sum_i \pi_i L_i(k)},
\end{equation}
and conditional source-count allocations $(k_1,k_2,k_3)$ summing to $k$ follow
\begin{equation}\label{eq:15}
\begin{split}
&\Prob(k_1,k_2,k_3 \mid j, K=k) \\
&\quad = \frac{\ind\{\sum_{b=1}^3 k_b = k\}}{L_j(k)} \prod_{b=1}^3 \binom{n_b}{k_b} q_{j,b}^{k_b} (1-q_{j,b})^{n_b-k_b}.
\end{split}
\end{equation}
Full feature vectors are sampled from the corresponding source-region clouds to construct conditional training sets, on which independent single-generation refits are trained from scratch. For each refit we measure clean and perturbed probabilities jointly, preserving their pairing when averaging transitions $\widehat{P}_t$ (Appendix~\ref{app:A}), and estimate $\widehat{h}_G^D(x)$ as the empirical fraction of refits failing $\tau_D(p)$. Between calibration anchors, convex interpolation combines response banks while~\eqref{eq:5} uses the exact target quotas and prevalence $p$. Crucially, all forecasts are frozen before any independent recursive validation chain is executed.

\emph{Telescoping contrast and baselines.} Comparing any two configurations $a$ and $b$ sharing initial law $\mu$ yields the telescoping decomposition
\begin{equation}\label{eq:16}
\begin{split}
R_b - R_a ={} & \mu \Bigl(\prod_{t<G} P_{b,t}\Bigr) (h_b - h_a) \\
& + \sum_{t<G} \mu \Bigl(\prod_{i<t} P_{b,i}\Bigr) (P_{b,t} - P_{a,t}) \Bigl(\prod_{i>t} P_{a,i}\Bigr) h_a,
\end{split}
\end{equation}
where each transition difference splits into a sampling-kernel shift and a learner-response shift:
\begin{equation}\label{eq:17}
P_b - P_a = \int (B_b - B_a)\,dF_a + \int B_b\,d(F_b - F_a).
\end{equation}
Equations~\eqref{eq:16}--\eqref{eq:17} directly motivate three structured ablations tested in our experiments: (1) the \emph{zero-perturbation response baseline}, which updates the sampling kernel $B_b$ to the target perturbation while keeping the unperturbed response bank $F_a, h_a$; (2) the \emph{conditional-mean transfer baseline}, which replaces paired response variability in $F_t$ with conditional means (losing tail variance; Appendix~\ref{app:C}); and (3) the \emph{proportion-copy baseline} in the image domain, which assumes clean outputs simply copy training-region counts $(K/N, C/N)$ mapped through a pilot $3\times 3$ perturbation table.

\emph{Error propagation and boundary resolution.} Let $\TV(a,b) = \frac{1}{2}\sum_y |a(y)-b(y)|$ denote total variation distance, define row-wise transition error $\epsilon_t(x) = \TV(P_t(x,\cdot), \widehat{P}_t(x,\cdot))$ and readout error $\epsilon_h(x) = |h_G^D(x) - \widehat{h}_G^D(x)|$, and set backward value vectors $\widehat{v}_G = \widehat{h}_G^D$ and $\widehat{v}_t = \widehat{P}_t \widehat{v}_{t+1}$ with oscillation $\osc(f) = \max_x f(x) - \min_x f(x)$.

\begin{theorem}[Propagation of calibration error]\label{thm:2}
Let $\widehat{R} = \widehat{\mu}\widehat{P}_0 \cdots \widehat{P}_{G-1}\widehat{h}_G^D$, where each $\widehat{P}_t$ is stochastic and $\widehat{h}_G^D \in [0,1]^{|\mathcal{S}|}$. With expectations under the true count marginals $\mu_t$,
\begin{equation}\label{eq:18}
\begin{split}
|R_G^D - \widehat{R}| \le{} & \TV(\mu,\widehat{\mu})\osc(\widehat{v}_0) \\
& + \sum_{t<G} \osc(\widehat{v}_{t+1})\E[\epsilon_t(X_t)] + \E[\epsilon_h(X_G)].
\end{split}
\end{equation}
\end{theorem}

\begin{proof}
Expanding $R_G^D - \widehat{R}$ with true prefixes $\mu_t = \mu P_0 \cdots P_{t-1}$ and approximate suffixes $\widehat{v}_t$ gives $(\mu-\widehat{\mu})\widehat{v}_0 + \sum_{t<G} \mu_t(P_t-\widehat{P}_t)\widehat{v}_{t+1} + \mu_G(h_G^D - \widehat{h}_G^D)$. For any probability vectors $a,b$ and bounded vector $f$, $(a-b)\mathbf{1}=0$ implies $|(a-b)f| = |(a-b)(f - \min f)| \le \TV(a,b)\osc(f)$. Applying this inequality term by term establishes~\eqref{eq:18}.
\end{proof}

Theorem~\ref{thm:2} shows that forecast error is governed strictly by calibration mismatch at count states actually visited by the true chain, weighted by the sensitivity $\osc(\widehat{v}_{t+1})$ of future risk to count perturbations (further decomposed into row-coupling and cutoff-proximity terms in Appendix~\ref{app:B}). Importantly, if a learner undergoes systematic multi-generation drift not represented in the single-generation bank, $\E[\epsilon_t(X_t)]$ and $\E[\epsilon_h(X_G)]$ grow directly---a mechanism we empirically verify in Section~\ref{sec:multilearner}.

Finally, on any one-dimensional continuous parameter cut $z$ where adjacent grid forecasts $\widehat{R}_i, \widehat{R}_{i+1}$ lie on opposite sides of budget $\beta$, linear interpolation locates the crossing display coordinate
\begin{equation}\label{eq:20}
\widehat{z}_* = z_i + (z_{i+1}-z_i)\frac{\beta - \widehat{R}_i}{\widehat{R}_{i+1} - \widehat{R}_i}.
\end{equation}

\begin{proposition}[Local boundary resolution]\label{prop:4}
Suppose $R(z)$ is differentiable on a cut $I$ with $R(z_*)=\beta$ and $0 < \lambda \le |R'(z)| \le L$. If two adjacent grid nodes spaced by $\Delta z$ satisfy $|\widehat{R}_k - R(z_k)| \le \varepsilon$ and bracket $\beta$, then~\eqref{eq:20} satisfies
\begin{equation}\label{eq:21}
|\widehat{z}_* - z_*| \le \frac{\varepsilon + L\Delta z / 2}{\lambda}.
\end{equation}
\end{proposition}

\begin{proof}
At interpolation weight $w \in [0,1]$, the linear interpolant of forecasts differs from that of true values by at most $\varepsilon$, while $L$-Lipschitz continuity bounds the gap between $R(z)$ and its nodal interpolant by $2w(1-w)L\Delta z \le L\Delta z/2$. Evaluating at $\widehat{z}_*$ gives $|R(\widehat{z}_*) - \beta| \le \varepsilon + L\Delta z/2$, and applying the mean value theorem with $|R'| \ge \lambda$ completes the proof.
\end{proof}

Proposition~\ref{prop:4} separates probability accuracy from boundary localization: even small probability error $\varepsilon$ translates into wide boundary uncertainty wherever the risk slope $\lambda$ is shallow, motivating our dual reporting of probability MAE and simultaneous empirical crossing brackets throughout the experiments.

\section{Controlled Count Verification and Two-Dimensional Diffusion Boundaries}\label{sec:count_ddpm}

We evaluate the forecasting engine along an escalating evidence ladder: first verifying the exact count law and geometric crossing reversal in controlled benchmarks with known output distributions, then testing frozen single-generation forecasts against independent recursive neural diffusion chains.

\subsection{Analytical Count Benchmark and Three-Axis Exactness}

On a controlled symmetric count learner outputting rare proportion $K/N$, exact finite-state propagation~\eqref{eq:13} reproduces large-scale Monte Carlo trajectories to within a tenth of a percentage point and brackets every empirical risk crossing. Across 2,601 settings ($8.52 \times 10^7$ simulated trajectories), exact propagation matches empirical six-round risk with 0.073~pp MAE, and all 123 empirically bracketed 20\% crossings contain the exact transition. (Exhaustive enumeration across 192 configurations independently verifies the discrete count-law convolution~\eqref{eq:5} and one-row substitution identity~\eqref{eq:9} to machine precision, with maximum discrepancies of $3.89 \times 10^{-16}$ and $4.09 \times 10^{-16}$.)

\begin{figure*}[!t]
\centering
\includegraphics[width=0.85\textwidth]{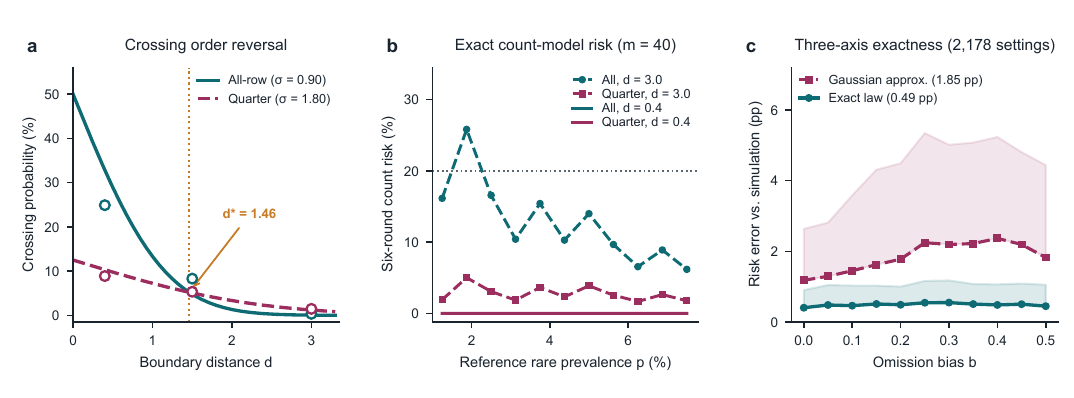}
\caption{Controlled count benchmark and three-axis exactness. (a) Analytical pointwise crossing probabilities (curves) and integrated truncated-Gaussian component probabilities (open circles at $d_0 \in \{0.4, 1.5, 3.0\}$) for equal-variance all-row ($\sigma=0.90$) and quarter-row ($\sigma=1.80$) policies, confirming the unique reversal at $d_* = 1.46$ (Proposition~\ref{prop:1}). (b) Exact six-round count-model terminal risk versus reference rare prevalence $p$ at $m=40$, illustrating reversal of policy risk ordering between near ($d_0=0.4$) and far ($d_0=3.0$) components and discrete sawtooth jumps as $\tau_A(p)$ crosses integer counts. (c) Three-axis $(\alpha, b, p)$ validation across 2,178 settings (4,096 trajectories per setting) varying synthetic share $\alpha$, rare-mode omission bias $b$, and prevalence $p$: exact multinomial propagation maintains 0.49~pp MAE (shaded band shows 90th percentile), whereas a Gaussian approximation degrades to 1.85~pp MAE as omission bias $b$ increases.}
\label{fig:3}
\end{figure*}

Shifting the mode separation across the unique threshold $d_* = 1.463$ reverses both single-step crossing probabilities and six-round terminal risk orderings (Proposition~\ref{prop:1}, Fig.~\ref{fig:3}a--b). For $\sigma=0.90$ and $a=2$, integrating over truncated Gaussian components ($\omega=0.65$) yields all-row versus quarter-row crossing probabilities of 24.90\% versus 8.88\% at $d_0=0.4$, reversing to 0.34\% versus 1.46\% at $d_0=3.0$. Propagating these exact rates through~\eqref{eq:13} at $m=40$ flips the terminal risk ordering across geometries and introduces non-monotone sawtooth jumps whenever the discrete threshold $K_G/N < p/2$ crosses an integer count.

Across a broader 2,178-setting three-axis grid varying synthetic share $\alpha$, rare-mode omission bias $b \in [0, 0.5]$, and prevalence $p$ (4,096 trajectories per setting), exact discrete propagation cuts forecast error nearly fourfold relative to a Gaussian approximation (Fig.~\ref{fig:3}c). Exact propagation achieves 0.49~pp MAE, places 2,061 of 2,178 forecasts inside empirical 95\% Wilson intervals, and matches the empirical side of the 20\% budget on 2,065 of 2,067 \emph{decided settings} (those whose 95\% interval lies entirely on one side of 20\%). By contrast, a Gaussian normal approximation suffers 1.85~pp MAE, covers only 927 of 2,178 intervals, and diverges systematically as omission bias grows---confirming that exact discrete tails are essential for rare-event forecasting.

\subsection{Two-Dimensional DDPM Protocol and Sequential Inference}

We next evaluate frozen forecasts on unconditionally trained denoising diffusion probabilistic models (DDPMs)~\cite{ref18}. The reference mixture $H_p$ places two truncated Gaussian components at $(-1.5, 0)$ and $(1.5, 0.7)$ with coordinate standard deviations $(0.65, 0.55)$, each restricted to its side of $x_1 = 0$. The rare region is the positive half-space $A = \{x_1 > 0\}$, and core $B \subset A$ is the ellipse centered at $(1.5, 0.7)$ with radii $(0.8, 0.8)$. At each round ($N=128$, $G=6$), a two-layer SiLU noise predictor (width 64) is trained from scratch for 1,500 Adam steps (learning rate $10^{-3}$) under a 100-step linear variance schedule ($10^{-4}$ to $0.15$) and sampled with all 100 ancestral reverse steps.

Calibration uses six sparse anchors ($p \in \{2.5\%, 7.5\%\}$, $m \in \{24, 48, 72\}$) with disjoint pilot chains (32 chains per anchor, 4,096 outputs saved at depths $0, 1, 3, 5$). We evaluate no perturbation, all-row perturbation at $\sigma \in \{0.45, 0.90\}$, quarter-row perturbation at $\sigma = 1.80$ (matching the all-row $\sigma=0.90$ variance $\eta\sigma^2 = 0.81$), and a held-out all-row intensity $\sigma = 0.675$ excluded from response fitting. The frozen all-row grid also includes an intermediate level $\sigma = 0.225$ that was not independently validated. Propagating scalar state $K_t$ with separate terminal readouts for $A$ and $B$ produces a frozen 10,404-entry grid over 17 prevalences and 17 real quotas before any validation chain is launched.

Validation proceeds in three disjoint stages (primary validation, targeted confirmation, and a low-$p$ endpoint check) with independent random streams. Each terminal generator's coverage is tested sequentially at cumulative sample sizes $8{,}192$, $32{,}768$, $131{,}072$, and $524{,}288$ using exact one-sided Clopper--Pearson limits~\cite{ref19} controlling misclassification error at $\delta = 0.0025$ (Appendix~\ref{app:D}). Definite and possible failure counts then yield conservative simultaneous 95\% chain-level risk intervals accounting for finite terminal sampling error.

\subsection{Frozen Forecasts and Operating Boundaries on DDPM Chains}

\begin{table}[!t]
\caption{Primary Two-Dimensional DDPM Validation Across 42 Settings per Endpoint. Errors Are in Percentage Points (pp); Right Columns Report Forecasts Inside Simultaneous 95\% Intervals and Correctly Classified Decided Settings.}
\label{tab:2}
\centering
\setlength{\tabcolsep}{3.2pt}
\footnotesize
\begin{tabular}{@{}lcccccc@{}}
\toprule
 & \multicolumn{2}{c}{MAE (pp)} & \multicolumn{2}{c}{Inside CI (/42)} & \multicolumn{2}{c}{Side match} \\
\cmidrule(lr){2-3}\cmidrule(lr){4-5}\cmidrule(lr){6-7}
Response model & $A$ & $B$ & $A$ & $B$ & $A$ (/33) & $B$ (/34) \\
\midrule
Perturbation-conditioned & 0.88 & \textbf{2.60} & 42 & \textbf{41} & 33 & \textbf{34} \\
Conditional-mean transfer & \textbf{0.65} & 2.63 & 42 & 40 & 33 & \textbf{34} \\
Zero-perturbation response & 1.31 & 13.03 & 39 & 14 & 33 & 24 \\
\bottomrule
\end{tabular}
\end{table}

Conditioning the single-generation response bank on the active perturbation cuts six-round core-risk MAE fivefold across the 42 primary validation settings (Table~\ref{tab:2}). Relative to the zero-perturbation baseline, calibrated error drops from 13.03~pp to 2.60~pp, simultaneous 95\% interval coverage rises from 14/42 to 41/42 settings, and all 34 decided core settings are classified on the correct side of 20\% (versus 24/34). A measurement-adjusted paired Brier difference of $-0.0282$ (95\% CI $[-0.0389, -0.0184]$) confirms this gain over the unperturbed baseline. On DDPM alone, calibrated pairing and conditional-mean transfer are statistically indistinguishable ($-0.00010$, $[-0.00059, 0.00038]$), though calibrated pairing proves significantly better on flow-matching, normalizing-flow, and KDE learners (Section~\ref{sec:multilearner}). On rare region $A$, risks stay low across most positive-noise settings and both calibrated models keep MAE below 0.9~pp.

At the \emph{focal equal-variance setting} ($m=40$, $p=5\%$, selected for paired confirmation), frozen forecasts closely track both primary validation and an independent confirmatory cohort. The calibrated model predicts six-round core risks of 25.63\% under all-row perturbation ($\sigma=0.90$) and 50.19\% under quarter-row perturbation ($\sigma=1.80$), closely matching primary estimates of 26.39\% and 52.43\%; at the held-out intensity $\sigma=0.675$, it forecasts 42.64\% against 44.27\% observed. Across 1,152 independent paired confirmatory chains, the measured all-row-minus-quarter contrast is $-26.74$~pp (95\% CI $[-36.04, -16.34]$~pp), containing the frozen forecast of $-24.56$~pp.

\begin{figure*}[!t]
\centering
\includegraphics[width=\textwidth]{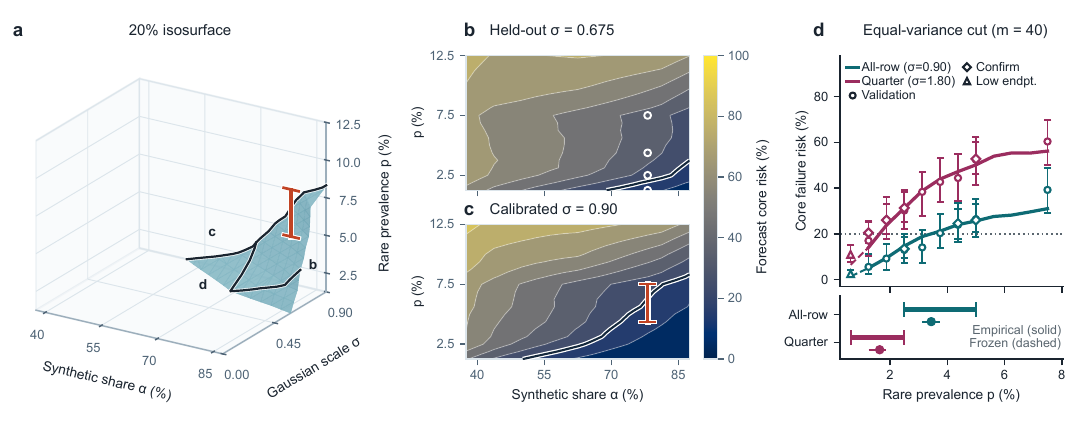}
\caption{Frozen two-dimensional DDPM core-risk forecasts and independent boundary validation. \textbf{(a)}~The 20\% isosurface is extracted within the $17 \times 5 \times 17$ frozen grid for synthetic fraction $\alpha$, Gaussian perturbation strength $\sigma$, and reference rare mass $p$. The mesh interpolates the numerical forecasts for display. Three dark slice curves mark the locations of panels~(b)--(d), and the orange-red vertical segment shows the independently validated crossing bracket at $\sigma = 0.90$, $m = 28$ ($\alpha = 78.125\%$). \textbf{(b, c)}~Two-dimensional forecast core-risk maps in $(\alpha, p)$ at held-out $\sigma = 0.675$ and calibrated $\sigma = 0.90$, showing dark 20\% contours with white halos. White open markers in~(b) mark confirmation settings at $m = 28$; the orange-red vertical segment in~(c) joins independently validated endpoints bracketing the 20\% crossing at $m = 28$ ($\alpha = 78.125\%$). \textbf{(d)}~Equal-variance core-failure risk curves at $m = 40$ ($\alpha = 68.75\%$) comparing all-row ($\sigma = 0.90$, teal) and quarter-row ($\sigma = 1.80$, plum) policies against validation, confirmation, and low-endpoint chains with joint 95\% intervals. The lower strip compares empirical crossing brackets (solid bars) against frozen grid crossings (dashed segments with linearly interpolated dots).}
\label{fig:4}
\end{figure*}

\begin{table}[!t]
\caption{Two-Dimensional DDPM 20\% Core-Risk Crossings in Reference Rare Prevalence $p$ (\%).}
\label{tab:3}
\centering
\setlength{\tabcolsep}{3.5pt}
\footnotesize
\begin{tabular}{@{}rlll@{}}
\toprule
$m$ & Policy & Frozen grid (\%) & Empirical bracket (\%) \\
\midrule
40 & All-row, $\sigma = 0.90$ & 3.125--3.750 & 2.500--5.000 \\
40 & Quarter-row, $\sigma = 1.80$ & 1.250--1.875 & 0.625--2.500 \\
28 & All-row, $\sigma = 0.90$ & 6.250--6.875 & 4.375--8.750$^\dagger$ \\
28 & All-row, $\sigma = 0.675$ (held-out) & 1.875--2.500 & Unresolved$^\ddagger$ \\
56 & All-row, $\sigma = 0.90$ & 1.250--1.875 & Unresolved \\
\bottomrule
\multicolumn{4}{@{}p{0.96\columnwidth}@{}}{\scriptsize $^\dagger$Confirmation slice-wise bracket; combining all stages tightens this to 4.375--7.500\%. $^\ddagger$Resolved on FM-Euler, FM-Heun, VAE, and KDE in Section~\ref{sec:multilearner}.}
\end{tabular}
\end{table}

Frozen 20\% operating boundaries fall inside independent empirical brackets wherever the risk slope is steep, while flat regimes remain unresolved as predicted by Proposition~\ref{prop:4} (Fig.~\ref{fig:4}, Table~\ref{tab:3}). Figure~\ref{fig:4}a visualizes the full three-dimensional operating boundary $\mathcal{B}^B$ defined in~\eqref{eq:boundary} across $(\alpha, \sigma, p)$, where the 20\% surface emerges above $\sigma \approx 0.5$ and passes through the independent crossing bracket at $m=28, \sigma=0.90$. Across the two-dimensional $(\alpha,p)$ slices (Fig.~\ref{fig:4}b--c), the calibrated $\sigma=0.90$ crossing at $m=28$ (frozen 6.250--6.875\%) lies inside its combined empirical bracket (4.375--7.500\%), whereas shallow risk slopes at $m=56$ and held-out $\sigma=0.675$ leave DDPM brackets unresolved. On the equal-variance cut at $m=40$ (Fig.~\ref{fig:4}d), both frozen grid crossings fall inside their simultaneous empirical brackets, predicting an all-row-minus-quarter crossing displacement of 1.80~pp in $p$ against an empirical bracket of $[0.000, 4.375]$~pp. A supplementary low-prevalence check ($p=0.625\%$, 1,536 chains) confirms that both policies lie well below 20\%, with observed core risks of 2.54\% ($[0.98\%, 4.28\%]$) for all-row and 11.00\% ($[7.53\%, 15.00\%]$) for quarter-row.

\section{Cross-Learner Generalization, Drift Limits, and Geometric Reversal}\label{sec:multilearner}

A forecasting method validated on one diffusion learner leaves two critical questions open: does single-generation calibration generalize across distinct generative model classes (and fail predictably when its stationarity assumption breaks), and does the geometric crossing reversal of Proposition~\ref{prop:1} persist from one step through six recursive rounds? We address both questions through a ten-learner benchmark and a controlled near/far geometry intervention.

\subsection{Generalization Across Ten Learners and Drift Diagnostics}

We replicated the full end-to-end protocol---pilot chains, six-anchor single-generation calibration, frozen forecast grids, and independent sequential validation---across ten learners spanning eight architectural classes: DDPM, standard DDIM~\cite{ref24}, truncated DDIM, flow matching~\cite{ref22} with Euler (FM-Euler) and Heun (FM-Heun) solvers, a generative adversarial network (GAN), an autoregressive density learner (AR), a normalizing flow (Norm.\ Flow), a variational autoencoder (VAE), and a kernel density estimator (KDE).

\begin{table}[!t]
\caption{Six-Round Core-Risk Forecast Accuracy and Multi-Round Stability Probe Across Ten Learners. MAE Is in Percentage Points (pp); Probe Reports $\log(c_5/c_3)$ and Round-6 Rare-Side Outlier Rate Under All-Row $\sigma=0.90$.}
\label{tab:4}
\centering
\setlength{\tabcolsep}{3.0pt}
\footnotesize
\begin{tabular}{@{}lrrrrrr@{}}
\toprule
 & & \multicolumn{2}{c}{Core MAE (pp)} & Side & & Outl. \\
\cmidrule(lr){3-4}
Learner & Settings & Cal. & Zero & match & $\log(c_5/c_3)$ & (\%) \\
\midrule
DDPM & 42 & \textbf{2.60} & 13.03 & 34/34 & $-0.025$ & 13.9 \\
Autoregressive$^\dagger$ & 53 & 3.27 & 16.93 & 22/23 & $-0.072$ & 21.9 \\
VAE$^\dagger$ & 47 & 3.30 & 9.45 & 34/34 & $-0.038$ & 15.5 \\
KDE$^\dagger$ & 42 & 3.82 & 15.57 & 15/17 & $-0.070$ & 24.4 \\
Norm.\ Flow$^\dagger$ & 46 & 4.29 & 16.76 & 32/32 & $-0.057$ & 21.4 \\
FM (Heun) & 42 & 4.56 & 18.10 & 31/31 & $-0.045$ & 17.5 \\
FM (Euler) & 42 & 4.85 & 17.73 & 30/30 & $-0.040$ & 16.3 \\
DDIM (trunc.) & 35 & 4.91 & 23.50 & 31/31 & $-0.047$ & 12.5 \\
GAN$^\dagger$ & 42 & 5.69 & 17.34 & 32/33 & $-0.064$ & 18.3 \\
\midrule
DDIM (std.) & 35 & 16.38 & 34.72 & 32/33 & $\mathbf{-0.195}$ & \textbf{47.4} \\
\bottomrule
\multicolumn{7}{@{}p{0.96\columnwidth}@{}}{\scriptsize $^\dagger$Stability probe evaluated prospectively prior to validation; the other five learners (DDPM, standard/truncated DDIM, FM-Euler, FM-Heun) were evaluated post hoc to diagnose DDIM drift and set the probe thresholds.}
\end{tabular}
\end{table}

\begin{figure*}[!t]
\centering
\includegraphics[width=0.85\textwidth]{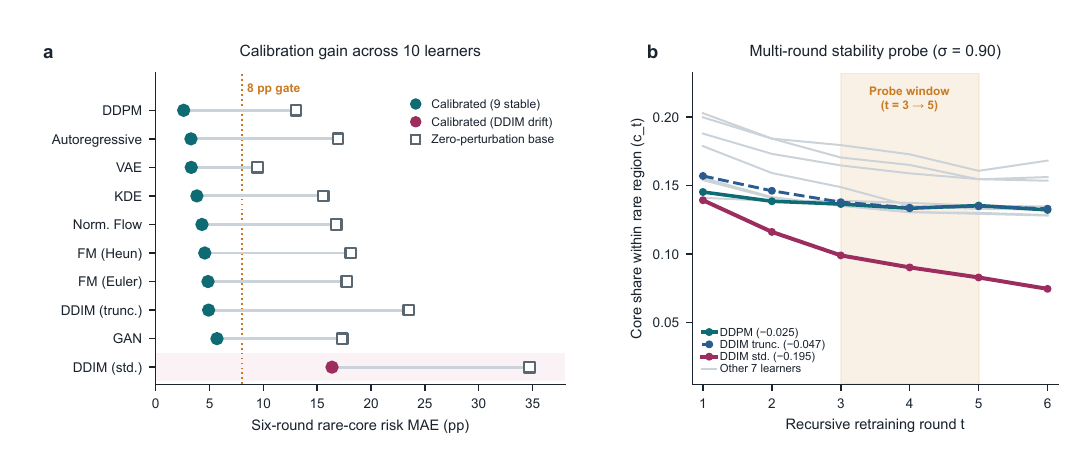}
\caption{Cross-learner generalization across ten learners and non-stationary drift diagnosis. (a) Paired comparison of six-round core-risk MAE between perturbation-conditioned calibration (filled circles) and the zero-perturbation response baseline (open squares). Nine stationary learners achieve 2.60--5.69~pp MAE (below the 8~pp diagnostic gate, amber dotted line), cutting error by 65--81\% relative to the unperturbed baseline, whereas standard DDIM fails at 16.38~pp (shaded band). (b) Multi-round stability probe tracking core share $c_t$ within rare region $A$ across rounds $t=1,\dots,6$ under all-row $\sigma=0.90$ (768 probe chains per learner; evaluated post hoc on the five diffusion/flow-matching learners to diagnose DDIM drift and prospectively prior to validation on GAN, AR, Norm.\ Flow, VAE, and KDE). Standard DDIM exhibits steep continuous erosion ($\log(c_5/c_3) = -0.195$, 47.4\% round-6 outliers), whereas truncating DDIM sampling ($\log(c_5/c_3) = -0.047$) restores stationarity and aligns with DDPM and the other seven learners.}
\label{fig:5}
\end{figure*}

Single-generation calibration generalizes reliably across nine of the ten learners, while cleanly failing on one non-stationary learner (Fig.~\ref{fig:5}a, Table~\ref{tab:4}). On the nine stationary learners, perturbation-conditioned calibration holds six-round core MAE to 2.60--5.69~pp versus 9.45--23.50~pp for the zero-perturbation baseline---a 65--81\% error reduction. Paired core Brier differences against the zero-perturbation baseline lie strictly below zero for all ten learners, ranging from $-0.0133$ (VAE) to $-0.1330$ (standard DDIM; $-0.0893$ among the nine stationary learners), and calibrated pairing also beats conditional-mean transfer on FM-Euler, FM-Heun, Norm.\ Flow, and KDE (Appendix~\ref{app:G}). In addition, four learners (FM-Euler, FM-Heun, VAE, and KDE) resolve empirical 20\% brackets at held-out $\sigma=0.675$ ($m=28$) where DDPM was inconclusive, and Norm.\ Flow strictly separates the $m=40$ equal-variance empirical brackets by $[0.625, 5.000]$~pp.

Standard DDIM provides an instructive falsification counterexample: its calibrated core MAE reaches 16.38~pp, it covers only 2 of 35 empirical intervals, and its focal paired forecast ($-26.96$~pp) falls outside the observed confirmation interval ($[-26.55, -12.94]$~pp). Theorem~\ref{thm:2} predicts that single-generation calibration must degrade whenever the target recursive chain drifts outside the stationary conditional distribution sampled by the pilot bank. To diagnose this mechanism directly, we evaluated a multi-round stability probe post hoc across the first five diffusion and flow-matching learners (DDPM, standard/truncated DDIM, FM-Euler, and FM-Heun) on 768 independent probe chains ($p=5\%$, $m=40$, four policies), computing the log-change $\log(c_5/c_3)$ in the fraction of samples in rare region $A$ that lie inside core $B$ between rounds 3 and 5 (Fig.~\ref{fig:5}b). Standard DDIM drops sharply across every generation ($\log(c_5/c_3) = -0.195$, 95\% CI $[-0.225, -0.165]$), accumulating a 47.4\% rare-side outlier rate by round 6. Truncating the deterministic DDIM sampler eliminates this outlier runaway (12.5\% outliers, $\log(c_5/c_3) = -0.047$) and immediately lowers calibrated core MAE from 16.38 to 4.91~pp with 31/31 decided settings classified correctly.

Because those first five post-hoc runs were used to set the stability thresholds (95\% interval inside $\pm 0.10$ predicts stability; outside $\pm 0.15$ predicts failure) and pass criterion ($\text{MAE} \le 8$~pp, 100\% decided settings matched, and focal paired forecast inside CI), agreement on them serves as diagnosis rather than predictive evidence. We therefore tested the locked probe prospectively prior to running validation on the last five learners (GAN, AR, Norm.\ Flow, VAE, KDE). All five learners stayed well below 6~pp MAE and met the focal paired-contrast criterion, yet under the strict 100\% classification gate, only two of the four prospective ``stable'' predictions (Norm.\ Flow and VAE) satisfied every condition; GAN (32/33 decided) and KDE (15/17 decided) missed on one to two borderline settings, while AR fell into the inconclusive probe band ($[-0.103, -0.041]$, 22/23 decided). Thus, the stability probe reliably diagnoses gross multi-generation drift such as standard DDIM, while remaining an imperfect prospective classifier for subtle borderline misses.

\subsection{Geometric Reversal from One Step to Six Rounds}

We next test Proposition~\ref{prop:1} directly by shifting the two mixture centers from the \emph{near} geometry $(\pm 1.5)$ to the \emph{far} geometry $(\pm 3.0)$ while keeping core $B$ centered on the rare peak, fixing $N=128$, $G=6$, and the two equal-variance policies ($\sigma=0.90$ all-row versus $\sigma=1.80$ quarter-row). Within each geometry, we build dedicated single-generation response banks and compare three nested count states of increasing resolution: $K$, $(K,C)$, and $(K,C,J)$, where $J$ counts training rows inside a concentric inner ellipse $I \subset B$ of radius $0.6\times$ that of $B$ (Appendix~\ref{app:F}).

\begin{figure*}[!t]
\centering
\includegraphics[width=0.85\textwidth]{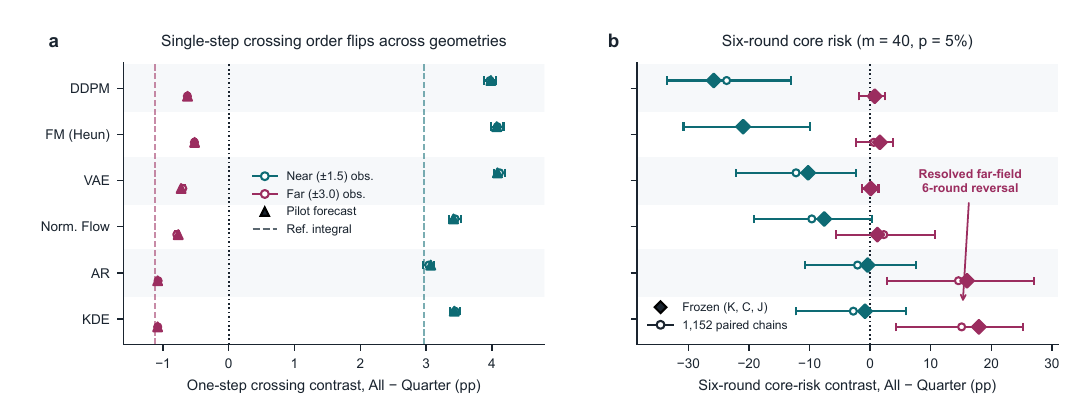}
\caption{Geometric intervention comparing near ($\pm 1.5$, teal) and far ($\pm 3.0$, plum) component separations across six representative learners at $p=5\%$. (a) Single-step boundary-crossing contrast across $x_1 = 0$ (all-row minus quarter-row, pp): reference integrals (dashed vertical lines: $+2.97$~pp near, $-1.12$~pp far), independent pilot forecasts (triangles), and observed contrasts over 384 fresh fitted parents per geometry (open circles with simultaneous 95\% intervals) flip sign cleanly across zero for every learner. (b) Six-round core-risk contrast at $m=40$ comparing frozen $(K,C,J)$ forecasts (diamonds) against 1,152 independent paired chains (open circles with simultaneous 95\% intervals). Whereas diffusion and VAE learners saturate above 98\% risk in the far geometry, non-saturating AR ($+14.58$~pp) and KDE ($+15.10$~pp) resolve the full six-round sign reversal predicted by the calibrated model.}
\label{fig:6}
\end{figure*}

In a single generation, \emph{all ten learners} confirm the sign reversal predicted by Proposition~\ref{prop:1}, matching independent pilot forecasts within tenths of a percentage point (Fig.~\ref{fig:6}a). Integrating Proposition~\ref{prop:1} over the truncated reference densities yields a crossing contrast (all-row minus quarter-row, counting boundary crossings in either direction per synthetic row) of $+2.97$~pp in the near geometry and $-1.12$~pp in the far geometry. Measured on 384 freshly fitted parents per geometry, every learner flips cleanly across zero: DDPM reverses from $+3.97$~pp ($[3.88, 4.07]$) to $-0.63$~pp ($[-0.66, -0.60]$), with comparable reversals on FM-Heun ($+4.08$ to $-0.52$~pp), AR ($+3.04$ to $-1.07$~pp), and KDE ($+3.44$ to $-1.08$~pp).

\begin{table*}[!t]
\caption{Effect of Count-State Richness on Far-Geometry Primary Core-Risk MAE (pp, 6 Settings) and Local 20\% Quota Crossings ($m$) at $p=15\%$ Using $(K,C,J)$.}
\label{tab:5}
\centering
\begin{tabular}{@{}lrrrll@{}}
\toprule
 & \multicolumn{3}{c}{Far-geometry primary MAE (pp)} & \multicolumn{2}{c}{Far-geometry $p=15\%$ crossing in real quota $m$} \\
\cmidrule(lr){2-4}\cmidrule(lr){5-6}
Learner & $K$ & $(K,C)$ & $(K,C,J)$ & All-row (frozen / empirical) & Quarter-row (frozen / empirical) \\
\midrule
DDPM & 3.67 & 2.41 & \textbf{2.23} & 104--116 / 104--124 & 104--116 / 92--124 \\
FM (Heun) & 4.92 & 3.19 & \textbf{3.04} & 116--124 / Unresolved & 124--128 / Unresolved \\
Norm.\ Flow & 7.26 & 3.53 & \textbf{3.09} & 104--116 / 92--124 & 104--116 / 92--116 \\
AR & 6.74 & 1.77 & \textbf{0.86} & 40--76 / 40--116 & 40--76 / 40--76 \\
KDE & 8.09 & 3.46 & \textbf{3.02} & 76--92 / 40--92 & 40--76 / 40--76 \\
VAE & 0.63 & 0.38 & \textbf{0.37} & None / Unresolved & None / Unresolved \\
\bottomrule
\end{tabular}
\end{table*}

Moving from one step to six recursive rounds uncovers two essential structural insights. First, finer count states consistently reduce far-geometry forecast error across every learner (Table~\ref{tab:5}): augmenting $K$ to $(K,C,J)$ cuts far-geometry core MAE from 3.67 to 2.23~pp on DDPM, from 7.26 to 3.09~pp on Norm.\ Flow, from 8.09 to 3.02~pp on KDE, and from 6.74 to 0.86~pp on AR. When components are widely separated, intra-core dispersion is no longer captured by the rare-region count $K$ alone.

Second, one-step transport reversal translates directly into a reversed six-round risk ordering whenever core failure does not saturate (Fig.~\ref{fig:6}b). In the near geometry ($m=40$, $p=5\%$), all-row perturbation yields markedly lower core failure risk than quarter-row perturbation across neural learners: DDPM shows a $-23.70$~pp contrast ($[-33.58, -13.04]$, forecast $-25.83$~pp), closely echoed by FM-Heun ($-20.92$~pp observed versus $-20.98$~pp forecast) and VAE ($-12.24$~pp versus $-10.25$~pp). In the far geometry at $m=40$, diffusion, flow-matching, GAN, and VAE learners all saturate above 98\% core failure under both policies, compressing their terminal risk contrast toward zero. Crucially, however, non-saturating AR and KDE operate in the interior of the risk curve (43--59\% failure) and directly resolve the predicted \emph{positive} six-round reversal: AR yields $+14.58$~pp ($[+2.84, +27.00]$, forecast $+15.98$~pp) and KDE yields $+15.10$~pp ($[+4.21, +25.25]$, forecast $+17.97$~pp). Sweeping real quota $m$ at $p=15\%$ in the far geometry further confirms that frozen $(K,C,J)$ crossings fall inside independent empirical brackets across DDPM, Norm.\ Flow, AR, and KDE (Table~\ref{tab:5}).

\section{Image-Space Validation on Fashion-MNIST}\label{sec:fmnist}

To test whether single-generation forecasting extends beyond low-dimensional Euclidean densities to high-dimensional image generation with learned representations, we study recursive training on Fashion-MNIST~\cite{ref20}.

\subsection{Image Protocol, No-Noise Accuracy, Efficiency, and Threshold Robustness}

We construct a two-class image mixture pairing T-shirt/top as the common class with Bag drawn at prevalence $p$. All generators train unconditionally on pixel images without class labels ($N=128$, $G=6$). Our primary image learner couples an eight-dimensional variational autoencoder (VAE)~\cite{ref21} with a latent flow-matching model~\cite{ref22} that generates latent codes; a plain VAE without latent flow serves as a high-failure control learner. A fixed classifier defines rare region $A$ as images predicted as Bag, and defines core $B \subset A$ as Bag-classified images whose Bag log-odds score meets or exceeds the median among correctly classified real Bags. This yields reference core prevalence $H_p(B) = 0.49767\,p$ ($H_{0.20}(B) = 0.09953$). Primary core failure uses $\tau_B(p) = \gamma H_p(B)$ (default $\gamma=1/2$), while the secondary rare-region endpoint uses nominal threshold $\tau_A(p) = \gamma p$ with real rows entering $A$ at exact classifier rate $H_p(A)$.

\begin{figure*}[!t]
\centering
\includegraphics[width=0.85\textwidth]{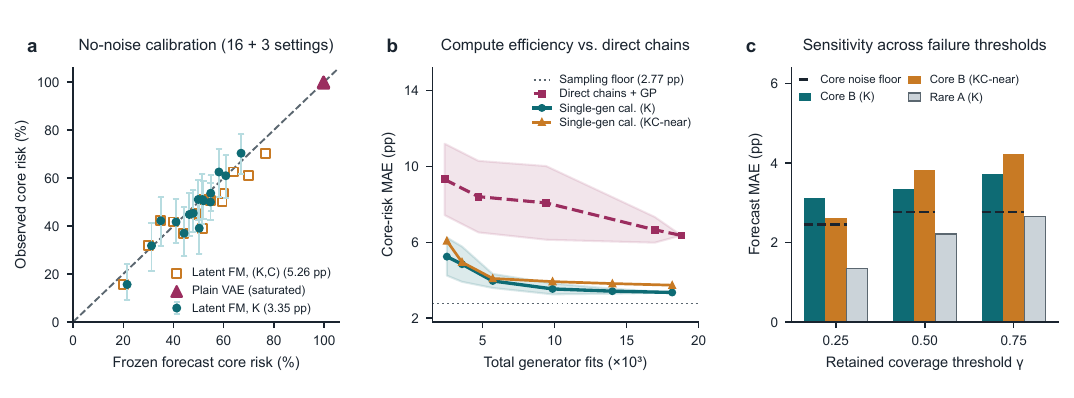}
\caption{Fashion-MNIST no-noise calibration, compute efficiency, and threshold robustness. (a) Observed versus frozen six-round core-failure risk across 16 latent flow-matching settings (192 independent chains each, $p \in [1.25\%, 12.5\%]$, $m \in [16, 76]$) and 3 plain-VAE saturation settings (triangles at 99.5--100\%). Scalar $K$-state calibration (teal circles, 3.35~pp MAE, 16/16 inside 95\% CI) tracks the identity line closely. (b) Calibration compute efficiency: single-generation calibration reaches 3.96~pp MAE at 5,694 fits and 3.54~pp at 9,852 fits (approaching the 2.77~pp finite-chain sampling floor, dotted line), outperforming leave-one-$p$-out Gaussian-process interpolation over direct recursive chains (6.36~pp even at 18,816 fits; shaded bands show $\pm 1$~SD across subsamples). (c) Robustness across retained coverage thresholds $\gamma \in \{0.25, 0.50, 0.75\}$: calibrated MAE closely tracks the core sampling noise floor (dashed segments) for both core $B$ and rare region $A$.}
\label{fig:7}
\end{figure*}

Without added perturbation, scalar $K$-state calibration forecasts six-round core risk nearly down to the finite-chain sampling floor across 16 latent flow-matching settings ($p \in [1.25\%, 12.5\%]$, $m \in [16, 76]$, 192 independent chains each; Fig.~\ref{fig:7}a). Scalar $K$-state propagation achieves 3.35~pp core MAE against the 2.77~pp sampling floor, places all 16 forecasts inside simultaneous 95\% intervals, and classifies all 15 decided settings correctly. Matching each state to the closest calibrated core count (\emph{nearest-$C$ lookup}) gives comparable accuracy (3.81~pp), whereas \emph{strict $(K,C)$ fallback}---backing off to scalar $K$ whenever subgroup counts are sparse---adds variance (5.26~pp). On rare region $A$, all three state variants stay within 0.3~pp of the 1.98~pp sampling floor, and on the plain-VAE control ($m=40$) frozen forecasts likewise capture complete saturation above 99.5\% core risk.

Two secondary analyses establish practical compute and threshold generality (Fig.~\ref{fig:7}b--c). First, single-generation $K$-state calibration consistently outperforms leave-one-$p$-out Gaussian-process (GP) interpolation over direct recursive chains, cutting out-of-sample MAE nearly in half at less than one-third the training-fit budget (3.96~pp at 5,694 fits versus 6.36~pp for direct-chain GP at 18,816 fits). Second, reusing the exact same single-generation response bank across stricter and looser failure definitions $\gamma \in \{0.25, 0.50, 0.75\}$ tracks the finite-chain sampling floor without retraining, keeping all 16 intervals covered and 100\% of decided settings aligned at every threshold.

\subsection{Latent-Perturbation Boundary Shift and Opposite Feedback Flows}

Because 15 of the 16 unperturbed settings above exceed the 20\% risk budget, we conducted a dedicated frozen boundary-displacement experiment at $p = 20\%$ comparing unperturbed feedback ($\sigma = 0$) against latent-space perturbation ($\sigma = 0.75$). Under $\sigma = 0.75$, each synthetic feedback image is decoded from $z' = z + \sigma a_M \odot \epsilon$, where $z$ is the parent's clean latent code, $a_M$ is the parent's aggregate posterior standard deviation vector, $\odot$ denotes element-wise multiplication, and $\epsilon \sim \mathcal{N}(0, I_8)$. Conditional responses are calibrated independently at six real quotas per policy and propagated via 65,536 nested $(K,C)$ trajectories (Appendix~\ref{app:F}).

\begin{table}[!t]
\caption{Frozen Six-Round Core Failure Forecasts and Independent Validation on Fashion-MNIST at $p = 20\%$ (Simultaneous 95\% Intervals Across All Four Settings).}
\label{tab:6}
\centering
\setlength{\tabcolsep}{3.5pt}
\footnotesize
\begin{tabular}{@{}rrrrrrr@{}}
\toprule
$\sigma$ & $m$ & $\alpha$ (\%) & Chains & Observed [95\% CI] & Cal. & Copy \\
\midrule
0.00 & 8 & 93.8 & 192 & 42.2 [32.6, 53.3] & 35.0 & 21.7 \\
0.00 & 72 & 43.8 & 576 & 8.9 [\phantom{0}5.9, 13.3] & 10.9 & 3.7 \\
0.75 & 72 & 43.8 & 192 & 35.4 [24.8, 45.9] & 34.6 & 21.0 \\
0.75 & 112 & 12.5 & 576 & 7.8 [\phantom{0}4.7, 12.3] & 10.3 & 4.7 \\
\midrule
\multicolumn{4}{@{}l}{Core MAE (pp) / Inside CI (/4)} & & \textbf{3.14} / \textbf{4} & 10.78 / 0 \\
\bottomrule
\end{tabular}
\end{table}

\begin{figure*}[!t]
\centering
\includegraphics[width=0.85\textwidth]{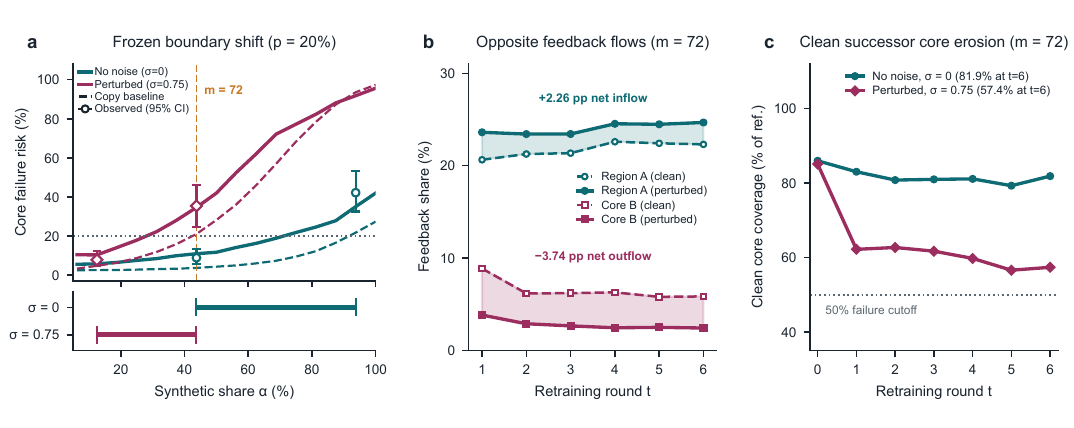}
\caption{Fashion-MNIST latent-perturbation boundary displacement and round-by-round opposite regional transport at $p = 20\%$. (a) Frozen calibrated core-risk curves (solid) versus proportion-copy baselines (dashed) across synthetic share $\alpha = 1 - m/128$ for $\sigma = 0$ (teal) and $\sigma = 0.75$ (plum), overlaid with independent validation observations and simultaneous 95\% intervals. Lower strip confirms non-overlapping empirical 20\% crossing brackets separated by the shared quota $m = 72$ ($\alpha = 43.75\%$, amber dashed line). (b) Round-by-round paired regional membership of the 56 synthetic feedback rows returned each round at $m=72, \sigma=0.75$: latent perturbation induces persistent net inflow into rare region $A$ ($21.8\% \to 24.0\%$, $+2.26$~pp average) while simultaneously causing persistent net outflow from core $B$ ($6.5\% \to 2.8\%$, $-3.74$~pp average). (c) Resulting clean successor core coverage relative to reference $H_p(B)$ across rounds $t=0,\dots,6$ at $m=72$: clean core retention drops to 57.4\% under $\sigma=0.75$ versus 81.9\% under $\sigma=0$.}
\label{fig:8}
\end{figure*}

Under latent perturbation at $p=20\%$, frozen calibration tracks observed core risk accurately and captures a decisive boundary shift across the shared quota $m=72$ ($\alpha=43.75\%$; Table~\ref{tab:6}, Fig.~\ref{fig:8}a). Independent chains place the 20\% crossing brackets on opposite sides of $m=72$, where enabling $\sigma=0.75$ drives observed core risk from below to strictly above 20\%. The frozen forecast is within 3.14~pp of the observed risk on average, and all four observations fall inside its simultaneous 95\% intervals (Table~\ref{tab:6}). Proportion copying is off by 10.78~pp and misses all four intervals, while alternative calibrated variants---nearest-$C$ lookup, a shared perturbation table, and scalar $K$-state---all remain within 1.27--3.53~pp MAE. Moreover, substituting the unperturbed response bank into the perturbed sampling law predicts only 8.79\% core risk at $(\sigma=0.75, m=72)$ against 35.42\% observed, confirming that perturbation-specific calibration remains indispensable in image space.

Finally, round-by-round paired flow logs at $m=72, \sigma=0.75$ directly verify the hidden-collapse mechanism sketched in Fig.~\ref{fig:1} (Stage~1). Across every round $t=1,\dots,6$, perturbing the 56 synthetic latent codes produces persistent net inflow into rare region $A$ ($+2.26$~pp/round) while simultaneously driving net outflow from core $B$ ($-3.74$~pp/round, with a 66.9--75.6\% per-round core exit rate; Fig.~\ref{fig:8}b). Fed back across six generations, this opposing transport drives clean core retention down to 57.4\% of reference versus 81.9\% without perturbation (Fig.~\ref{fig:8}c), raising clean classifier-feature Fr\'echet distance from 12.3 to 20.5 even as rare region $A$ remains above 20\% throughout.

\section{Discussion}\label{sec:discussion}

Across controlled count dynamics, ten two-dimensional learners, and Fashion-MNIST latent flow-matching chains, decoupling exact three-source count transport from single-generation conditional calibration forecasts finite-horizon core failure within roughly 2--5~pp MAE at a fraction of the computational cost of exhaustive recursive chain sweeps. Synthesizing these three evidence regimes clarifies the mechanism of hidden rare-mode collapse, positions our results relative to prior collapse theory, and defines sharp operational boundaries for when one-step calibration succeeds or fails.

\subsection{Why Coarse Monitoring Misses Core Erosion}

The recurring pattern across both Euclidean and learned image manifolds is that occupancy of rare region $A$ and retention of core $B$ can move in opposite directions under the same intervention. In two-dimensional Gaussian mixtures, outward spreading moves majority mass across $x_1 = 0$ while thinning density near the rare peak (Eq.~\eqref{eq:3}). On Fashion-MNIST, latent-space perturbation produces the exact same asymmetry through a learned decoder: leaked majority examples expand rare region $A$ every round while core $B$ prototypes disperse into periphery $A \setminus B$ (Fig.~\ref{fig:8}b). Because the paired accounting identity~\eqref{eq:2} depends only on regional transition probabilities rather than ambient geometry, it unifies both phenomena. Practically, this indicates that monitoring class-level proportions or coarse classifier rates during recursive deployment can systematically overlook severe loss of rare-mode fidelity.

\subsection{Positioning Within Recursive Training Theory}

Our findings extend prior theoretical accounts of model collapse in two concrete directions. First, whereas asymptotic stability criteria~\cite{ref6,ref8} and data-accumulation analyses~\cite{ref9,ref10,ref11} determine whether collapse eventually occurs as $G \to \infty$, practical training runs operate at finite horizons under fixed real-data budgets. Theorem~\ref{thm:1} and the three-axis count comparison (Fig.~\ref{fig:3}c) show that exact finite-sample count propagation captures discrete tail behavior that Gaussian approximations miss, while empirical calibration replaces intractable neural score-error constants~\cite{ref13,ref14} with directly measurable one-step responses. Second, the equal-variance geometry intervention demonstrates that feedback variance alone does not determine recursive stability: matching per-row variance $\eta\sigma^2$ can flip both one-step transport and six-round core-failure risk depending on mode separation relative to $d_*$ (Proposition~\ref{prop:1}, Fig.~\ref{fig:6}). This complements recent work on selection-induced tail loss~\cite{ref15,ref17} by showing that even value-independent perturbations induce geometry-dependent mode selection.

\subsection{When Single-Generation Calibration Holds---and Where It Breaks}

Taken together, the ten-learner benchmark and Theorem~\ref{thm:2} clarify the exact operating envelope of single-generation forecasting. When a learner's conditional response distribution given the tracked count state remains nearly stationary across rounds, one-step calibration captures the dominant nonlinearity of neural fitting, explaining why perturbation-conditioned models reduce core MAE by 65--81\% relative to unperturbed baselines across nine stationary learners and outperform leave-one-$p$-out direct-chain GP interpolation on Fashion-MNIST (Fig.~\ref{fig:7}b).

Two failure mechanisms bound this accuracy, both directly anticipated by our theory:
\begin{itemize}
\item \emph{Non-stationary multi-generation drift.} Standard DDIM violates the stationarity assumption because deterministic reverse trajectories accumulate off-manifold outliers across rounds (47.4\% by round 6), causing $\E[\epsilon_t(X_t)]$ and $\E[\epsilon_h(X_G)]$ in Theorem~\ref{thm:2} to grow well beyond single-generation estimates. Truncating the reverse sampler removes this drift and immediately restores calibration accuracy (16.38 to 4.91~pp MAE), confirming that the breakdown stems from sampler-induced non-stationarity rather than count-law misspecification.
\item \emph{Insufficient count-state granularity.} In the far two-dimensional geometry, scalar $K$ overlooks intra-core dispersion, inflating error up to 8.1~pp; enriching the state to nested counts $(K,C,J)$ cuts far-geometry MAE across every learner (down to 0.86--3.09~pp, Table~\ref{tab:5}). Conversely, on unperturbed Fashion-MNIST, sparse subgroup counts in $(K,C)$ slightly favor the lower-variance scalar $K$-state (3.35 versus 5.26~pp MAE), illustrating a classical bias--variance trade-off in state selection.
\end{itemize}

\subsection{Claim-Specific Boundaries and Discriminating Next Tests}

Four explicit boundaries qualify the transportability of our conclusions without overturning the within-design findings:
\begin{enumerate}
\item \emph{Cross-perturbation extrapolation.} Calibrated response banks are perturbation-specific: transferring a zero-perturbation bank to $\sigma=0.75$ on Fashion-MNIST underpredicts core risk by 26.6~pp (8.79\% versus 35.42\%). Applying the framework to a new corruption mechanism therefore requires re-calibrating one-step responses under that mechanism, though the count law~\eqref{eq:5} itself remains unchanged.
\item \emph{Image-domain breadth and decision-side discrimination.} The completed Fashion-MNIST perturbation study evaluates one class pair (T-shirt/Bag), one reference prevalence ($p=20\%$), and two latent noise levels ($\sigma \in \{0, 0.75\}$). Moreover, at the four validated settings, both the calibrated model and the proportion-copy baseline classify the same side of the 20\% budget, differing strictly in probability accuracy (Table~\ref{tab:6}). Testing intermediate noise levels ($\sigma \in \{0.25, 0.50\}$), additional image class pairs with distinct geometric separation, and quotas in the intermediate window $m \in [48, 64]$ where calibrated and copy forecasts disagree on the 20\% side provides the direct discriminating experiment for decision-level superiority.
\item \emph{Boundary localization under flat risk slopes.} As proven in Proposition~\ref{prop:4}, low probability MAE does not guarantee narrow empirical crossing brackets when local risk slopes are shallow (e.g., DDPM at $m=56$ or held-out $\sigma=0.675$). Tightening boundary brackets in flat regimes requires adaptive chain allocation concentrated near $\widehat{z}_*$ rather than uniform grid sweeps.
\item \emph{Prospective stationarity screening.} While the post-hoc stability probe $\log(c_5/c_3)$ cleanly separates catastrophic drift (standard DDIM) from stationary learners, prospective screening prior to validation on the five held-out learners showed that two of four ``stable'' calls still missed one to two borderline settings. Designing a multi-metric probe that monitors both core share and tail-outlier mass before committing calibration budgets is a concrete next step for automating deployment safety checks.
\end{enumerate}

\section{Conclusion}\label{sec:conclusion}

We introduced a decoupled single-generation forecasting framework for predicting finite-horizon core collapse and locating operating risk boundaries in recursive generative training. By pairing an exact three-source multinomial count-transport law with independent single-generation conditional learner fits, the framework predicts six-round core failure probabilities accurately across an exact three-axis count benchmark, nine stationary two-dimensional learners, and Fashion-MNIST latent flow-matching chains, while cutting training-fit budgets relative to direct multi-generation simulation. Together, the continuous flux and equal-variance reversal results explain why counts in rare region $A$ can expand even as core $B$ erodes, and the standard DDIM counterexample defines non-stationary drift as the primary failure mode of one-step calibration.

\section*{Code and Data Availability}

All experiment code, frozen forecast tables, chain-level validation summaries, paired flow logs, and figure-generation scripts required to reproduce Tables~\ref{tab:2}--\ref{tab:7} and Figs.~\ref{fig:1}--\ref{fig:8} will be made publicly available upon publication. Re-running neural training from scratch additionally requires the Fashion-MNIST dataset and the fixed classifier weights documented with the code release.

\appendices
\section{Matrix Approximation and Truncation Brackets}\label{app:A}

Write $f_{\mathrm{bin}}(j;n,q)$ for the binomial mass at $j$ and $*$ for discrete convolution. At a calibration anchor with $L_k$ independent refits at scalar count $k$, let $(q_\ell, q_\ell^\sigma)$ be the jointly measured clean and perturbed rare-region probabilities from refit $\ell$. The empirical transition row averages the per-refit convolutions:
\begin{equation}\label{eq:22}
\begin{split}
\widehat{P}(k,j) = L_k^{-1} \sum_{\ell=1}^{L_k} \bigl[ & f_{\mathrm{bin}}(\cdot; m, H_p(A)) * {} \\
& f_{\mathrm{bin}}(\cdot; s-u, q_\ell) * f_{\mathrm{bin}}(\cdot; u, q_\ell^\sigma) \bigr](j),
\end{split}
\end{equation}
preserving the pairing between $q_\ell$ and $q_\ell^\sigma$. The terminal readout for region $D$ is estimated by $\widehat{h}_G^D(k) = L_k^{-1}\sum_{\ell=1}^{L_k} \ind\{q_{\ell,D} < \tau_D(p)\}$. Between anchors, piecewise-linear convex weights in $(p, m, \sigma)$ combine response banks, with nearest-anchor clamping at domain boundaries and a dedicated bank for the quarter-row policy. Generation zero uses real-only conditional refits with initial law $\Bin(N, H_p(A))$; subsequent rounds use the pooled-depth bank.

In the two-dimensional diffusion calculation, count states are truncated to $\mathcal{S}_0 = \{0,\dots,48\}$. Letting $\widetilde{\mu}_0$ and $\widetilde{P}_t$ denote unrenormalized restrictions to $\mathcal{S}_0$, propagating $\widetilde{\mu}_{t+1} = \widetilde{\mu}_t \widetilde{P}_t$ tracks probability mass remaining inside $\mathcal{S}_0$ through round $G$. With lower bound $L = \widetilde{\mu}_G \widehat{h}_G^D$ and omitted mass $b = 1 - \widetilde{\mu}_G \mathbf{1}$, every stochastic completion of $\widetilde{P}_t$ produces a forecast inside the exact truncation bracket $[L, L+b]$. The reported point forecast directs overflow into state 48 after each step.

\section{Coupling Bounds and Smooth Boundary Sensitivity}\label{app:B}

To bound the transition error $\epsilon_t(x)$ in Theorem~\ref{thm:2} without comparing full laws of fitted models, couple a true parent pair $(g,\gs)$ with a calibrated pair $(\widehat{g},\widehat{g}^\sigma)$ at fixed state $x$, setting $d_0 = \TV(g,\widehat{g})$ and $d_1 = \TV(\gs,\widehat{g}^\sigma)$. Assigning identical fresh rows and maximally coupling synthetic row assignments~\cite{ref23} ensures identical next counts whenever all synthetic row assignments match, yielding
\begin{equation}\label{eq:23}
\begin{split}
\epsilon_t(x) &\le \E\bigl[1 - (1-d_0)^{s-u}(1-d_1)^u \mid x\bigr] \\
&\le \E\bigl[\min\{1, (s-u)d_0 + u d_1\} \mid x\bigr].
\end{split}
\end{equation}
Because equal full counts imply equal projections,~\eqref{eq:23} also bounds coarse-state transitions.

For terminal readout error, coupling true and calibrated clean coverages $(Q, \widehat{Q})$ conditional on $x$ shows that for any margin $a > 0$,
\begin{equation}\label{eq:24}
\epsilon_h(x) \le \Prob(|Q - \tau_D(p)| \le a \mid x) + \Prob(|Q - \widehat{Q}| > a \mid x),
\end{equation}
separating cutoff-proximity mass from response discrepancy. Furthermore, defining the Dobrushin contraction coefficient $c(P) = \max_{x,x'} \TV(P(x,\cdot), P(x',\cdot))$ gives $\osc(Pf) \le c(P)\osc(f)$ and inductively $\osc(\widehat{v}_{t+1}) \le \osc(\widehat{h}_G^D)\prod_{i=t+1}^{G-1} c(\widehat{P}_i)$.

Along any smooth protocol parameter $z$ where $\mu(z)$, $P_t(z)$, and $h_G^D(z)$ are continuously differentiable, differentiating~\eqref{eq:13} with $v_G = h_G^D$ and $v_t = P_t v_{t+1}$ gives
\begin{equation}\label{eq:25}
\frac{dR_G^D}{dz} = \mu' v_0 + \sum_{t<G} \mu_t P_t' v_{t+1} + \mu_G (h_G^D)'.
\end{equation}
Whenever $\partial_p R_G^D \neq 0$ at a root $(p_*(\sigma), \sigma)$ of $R_G^D(p,\sigma) = \beta$, the implicit function theorem yields the local boundary slope
\begin{equation}\label{eq:26}
\frac{dp_*}{d\sigma} = -\left.\frac{\partial_\sigma R_G^D}{\partial_p R_G^D}\right|_{(p_*(\sigma),\sigma)}.
\end{equation}

\section{Variance and Tail Loss Under Conditional-Mean Transfer}\label{app:C}

Under zero perturbation, let $Q$ denote the random parent rare-region probability conditional on training count $K_t$, with conditional mean $\bar{q} = \E[Q \mid K_t]$. By the law of total variance,
\begin{equation}\label{eq:27}
\begin{split}
\Var(K_{t+1} \mid K_t) ={} & mp(1-p) + s\bar{q}(1-\bar{q}) \\
& + s(s-1)\Var(Q \mid K_t),
\end{split}
\end{equation}
and by Jensen's inequality for $s \ge 1$,
\begin{equation}\label{eq:28}
\begin{split}
\Prob(K_{t+1}=0 \mid K_t) &= (1-p)^m \E[(1-Q)^s \mid K_t] \\
&\ge (1-p)^m (1-\bar{q})^s.
\end{split}
\end{equation}
Replacing $Q$ with $\bar{q}$ discards the overdispersion term $s(s-1)\Var(Q\mid K_t)$ and underestimates extreme extinction tails; under positive perturbation, replacing $(Q, Q^\sigma)$ with marginal means also discards their joint parent-level dependence.

\section{Sequential Terminal Evaluation and Simultaneous Risk Intervals}\label{app:D}

Let $Y_i$ denote the true terminal failure indicator of chain $i$. At predetermined cumulative evaluation sizes, a fitted generator is classified as definitely failing ($L_i=1$) when its upper one-sided Clopper--Pearson coverage limit lies below $\tau_D(p)$, and definitely passing ($U_i=0$) when its lower limit lies above $\tau_D(p)$; unresolved chains retain $(L_i, U_i) = (0,1)$. Allocating one-sided error budget $\delta = 0.0025$ across looks controls misclassification conditional on each fitted generator, yielding
\begin{equation}\label{eq:29}
\E[L_i] - \delta \le R_G^D \le \E[U_i] + \delta.
\end{equation}
Applying exact binomial limits to the chain totals of $L_i$ and $U_i$, shifting by $\mp \delta$, and clipping to $[0,1]$ yields conservative intervals that account for finite terminal sampling error. Bonferroni allocation across endpoints, stages (budgets $0.0125$, $0.025$, $0.0125$ in the three-stage equal-variance design), and directions ensures simultaneous 95\% coverage without assuming independence across paired policy arms. Given simultaneous crossing brackets $[a,b]$ and $[c,d]$ for two policies, the implied displacement bracket is $[a-d, b-c]$. Paired risk differences $\Prob(Y_a=1, Y_b=0) - \Prob(Y_a=0, Y_b=1)$ combine conservative bounds for both discordant cells.

\section{Controlled Count Model Crossing Integral}\label{app:E}

For the controlled symmetric count learner with $q = k/N$ and $q^\sigma = q + \ell(1-2q)$, a component centered at distance $d_0$ from $x_1 = 0$ with pre-truncation horizontal standard deviation $\omega$ has one-direction crossing probability
\begin{equation}\label{eq:30}
\ell(d_0, \sigma) = \frac{\int_0^\infty \Phibar(x/\sigma)\,\phi((x-d_0)/\omega)\,dx}{\omega\,\Phi(d_0/\omega)},
\end{equation}
with $\ell(d_0, 0) = 0$. Direct Monte Carlo simulation of source counts and bidirectional row crossings matches the exact Markov matrix risk to 0.073~pp MAE across 2,601 grid settings and 0.076~pp MAE across a separate 405-setting check ($1.33 \times 10^7$ trajectories).

\section{Nested Regional Counts and Subgroup Backoff}\label{app:F}

For nested regions $I \subset B \subset A$ with source probabilities $(a, b, c)$, a source of size $n$ generates exact nested counts via
\begin{equation}\label{eq:31}
\begin{aligned}
K &\sim \Bin(n, a), \\
C \mid K &\sim \Bin(K, b/a), \\
J \mid C &\sim \Bin(C, c/b),
\end{aligned}
\end{equation}
with zero subsequent counts whenever a conditioning denominator vanishes. Summing the three independent source vectors reproduces the exact multinomial law~\eqref{eq:5} on the four-region partition (verified against direct enumeration to $1.94 \times 10^{-16}$). When evaluating finer count states $(K,C)$ or $(K,C,J)$ from a response bank indexed by $K$, each of 65,536 forecast particles draws from a finer subgroup containing $n_x$ refits with backoff weight $n_x/(n_x+8)$ and otherwise backs off to the next coarser state, capping lookup counts at 48.

\section{Paired Brier Scores Across Ten Learners}\label{app:G}

Table~\ref{tab:7} reports measurement-adjusted paired core Brier differences (calibrated minus baseline; negative favors perturbation-conditioned calibration) across all ten two-dimensional learners evaluated in Section~\ref{sec:multilearner}. Calibration strictly outperforms the zero-perturbation baseline across all ten learners, and strictly outperforms conditional-mean transfer on DDIM (std.), FM-Euler, FM-Heun, Norm.\ Flow, and KDE.

\begin{table}[!t]
\caption{Paired Six-Round Core Brier Score Differences (Calibrated Minus Baseline; Negative Favors Calibration) With Measurement-Adjusted 95\% Intervals.}
\label{tab:7}
\centering
\setlength{\tabcolsep}{2.2pt}
\scriptsize
\begin{tabular}{@{}lrr@{}}
\toprule
Learner & vs.\ Zero-pert.\ [95\% CI] & vs.\ Cond.\ mean [95\% CI] \\
\midrule
DDPM & $-0.0282$ $[-0.0389, -0.0184]$ & $-0.00010$ $[-0.00059, \phantom{-}0.00038]$ \\
DDIM (std.) & $-0.1330$ $[-0.1440, -0.1206]$ & $-0.00159$ $[-0.00222, -0.00105]$ \\
DDIM (trunc.) & $-0.0893$ $[-0.1037, -0.0745]$ & $-0.00052$ $[-0.00104, \phantom{-}0.00002]$ \\
FM (Euler) & $-0.0460$ $[-0.0572, -0.0338]$ & $-0.00063$ $[-0.00111, -0.00011]$ \\
FM (Heun) & $-0.0466$ $[-0.0598, -0.0348]$ & $-0.00066$ $[-0.00124, -0.00014]$ \\
GAN & $-0.0559$ $[-0.0627, -0.0490]$ & $-0.00052$ $[-0.00122, \phantom{-}0.00017]$ \\
AR & $-0.0387$ $[-0.0474, -0.0309]$ & $-0.00008$ $[-0.00018, \phantom{-}0.00001]$ \\
Norm.\ Flow & $-0.0373$ $[-0.0449, -0.0296]$ & $-0.00041$ $[-0.00077, -0.00007]$ \\
VAE & $-0.0133$ $[-0.0197, -0.0071]$ & $-0.00002$ $[-0.00040, \phantom{-}0.00034]$ \\
KDE & $-0.0291$ $[-0.0370, -0.0219]$ & $-0.00014$ $[-0.00029, -0.00001]$ \\
\bottomrule
\end{tabular}
\end{table}


\begin{thebibliography}{24}

\bibitem{ref1}
I.~Shumailov, Z.~Shumaylov, Y.~Zhao, N.~Papernot, R.~Anderson, and Y.~Gal, ``AI models collapse when trained on recursively generated data,'' \emph{Nature}, vol. 631, pp. 755--759, 2024.

\bibitem{ref2}
A.~T. Suresh, A.~Thangaraj, and A.~N.~K. Khandavally, ``Rate of model collapse in recursive training,'' in \emph{Proc. Int. Conf. Artif. Intell. Stat. (AISTATS)}, ser. Proc. Mach. Learn. Res., vol. 258, 2025, pp. 1396--1404.

\bibitem{ref3}
S.~Wyllie, I.~Shumailov, and N.~Papernot, ``Fairness feedback loops: Training on synthetic data amplifies bias,'' in \emph{Proc. ACM Conf. Fairness, Accountability, and Transparency (FAccT)}, 2024, pp. 2113--2147.

\bibitem{ref4}
M.~S.~M. Sajjadi, O.~Bachem, M.~Lu\v{c}i\'{c}, O.~Bousquet, and S.~Gelly, ``Assessing generative models via precision and recall,'' in \emph{Adv. Neural Inf. Process. Syst.}, vol.~31, 2018.

\bibitem{ref5}
T.~Kynk\"{a}\"{a}nniemi, T.~Karras, S.~Laine, J.~Lehtinen, and T.~Aila, ``Improved precision and recall metric for assessing generative models,'' in \emph{Adv. Neural Inf. Process. Syst.}, vol.~32, 2019.

\bibitem{ref6}
Q.~Bertrand, J.~Bose, A.~Duplessis, M.~Jiralerspong, and G.~Gidel, ``On the stability of iterative retraining of generative models on their own data,'' in \emph{Proc. Int. Conf. Learn. Represent. (ICLR)}, 2024.

\bibitem{ref7}
M.~E.~A. Seddik, S.-W. Chen, S.~Hayou, P.~Youssef, and M.~Debbah, ``How bad is training on synthetic data? a statistical analysis of language model collapse,'' in \emph{Proc. Conf. Lang. Model. (COLM)}, 2024. [Online]. Available: \url{https://openreview.net/forum?id=t3z6UlV09o}

\bibitem{ref8}
S.~Alemohammad, J.~Casco-Rodriguez, L.~Luzi, A.~I. Humayun, H.~Babaei, D.~LeJeune, A.~Siahkoohi, and R.~Baraniuk, ``Self-consuming generative models go MAD,'' in \emph{Proc. Int. Conf. Learn. Represent. (ICLR)}, 2024.

\bibitem{ref9}
M.~Gerstgrasser, R.~Schaeffer, A.~Dey, R.~Rafailov, H.~Sleight, J.~Hughes, T.~Korbak, R.~Agrawal, D.~Pai, A.~Gromov, D.~A. Roberts, D.~Yang, D.~L. Donoho, and S.~Koyejo, ``Is model collapse inevitable? breaking the curse of recursion by accumulating real and synthetic data,'' in \emph{Proc. Conf. Lang. Model. (COLM)}, 2024.

\bibitem{ref10}
J.~Kazdan, R.~Schaeffer, A.~Dey, M.~Gerstgrasser, R.~Rafailov, D.~L. Donoho, and S.~Koyejo, ``Collapse or thrive: Perils and promises of synthetic data in a self-generating world,'' in \emph{Proc. Int. Conf. Mach. Learn. (ICML)}, ser. Proc. Mach. Learn. Res., vol. 267, 2025, pp. 29\,469--29\,494.

\bibitem{ref11}
R.~Turgeman and T.~Tirer, ``Model collapse is avoidable even as real data vanishes: A sharp criterion for iterative gaussian estimation,'' arXiv:2611.00259, 2026.

\bibitem{ref12}
H.~Malet and G.~Turinici, ``Feature selective model collapse in diffusion models: Total replacement versus fixed-budget training,'' arXiv:2610.01318, 2026.

\bibitem{ref13}
N.~B. Khelifa, R.~E. Turner, and R.~Venkataramanan, ``Quantifying error propagation and model collapse in diffusion models,'' in \emph{Proc. Int. Conf. Mach. Learn. (ICML)}, ser. Proc. Mach. Learn. Res., vol. 306, 2026, pp. 56\,941--56\,981.

\bibitem{ref14}
\rule{2em}{0.4pt}, ``Recursively trained diffusion models: Limiting collapse distribution and spectral characterization,'' arXiv:2606.13796v2, 2026.

\bibitem{ref15}
E.~Dohmatob, Y.~Feng, P.~Yang, F.~Charton, and J.~Kempe, ``A tale of tails: Model collapse as a change of scaling laws,'' in \emph{Proc. Int. Conf. Mach. Learn. (ICML)}, ser. Proc. Mach. Learn. Res., vol. 235, 2024, pp. 11\,165--11\,197.

\bibitem{ref16}
L.~Shi, M.~Wu, H.~Zhang, Z.~Zhang, M.~Tao, and Q.~Qu, ``A closer look at model collapse: From a generalization-to-memorization perspective,'' in \emph{Adv. Neural Inf. Process. Syst.}, vol.~38, 2025.

\bibitem{ref17}
X.~Qiao, X.~Du, W.~Liu, J.~Zhang, P.~Mai, M.~Zhang, and Y.~Pang, ``When sample selection bias precipitates model collapse,'' in \emph{Proc. Int. Conf. Mach. Learn. (ICML)}, ser. Proc. Mach. Learn. Res., vol. 306, 2026, pp. 101\,222--101\,268.

\bibitem{ref18}
J.~Ho, A.~Jain, and P.~Abbeel, ``Denoising diffusion probabilistic models,'' in \emph{Adv. Neural Inf. Process. Syst.}, vol.~33, 2020, pp. 6840--6851.

\bibitem{ref19}
C.~J. Clopper and E.~S. Pearson, ``The use of confidence or fiducial limits illustrated in the case of the binomial,'' \emph{Biometrika}, vol.~26, no.~4, pp. 404--413, 1934.

\bibitem{ref20}
H.~Xiao, K.~Rasul, and R.~Vollgraf, ``Fashion-MNIST: A novel image dataset for benchmarking machine learning algorithms,'' arXiv:1708.07747, 2017.

\bibitem{ref21}
D.~P. Kingma and M.~Welling, ``Auto-encoding variational bayes,'' in \emph{International Conference on Learning Representations}, 2014.

\bibitem{ref22}
Y.~Lipman, R.~T.~Q. Chen, H.~Ben-Hamu, M.~Nickel, and M.~Le, ``Flow matching for generative modeling,'' in \emph{International Conference on Learning Representations}, 2023.

\bibitem{ref23}
D.~A. Levin and Y.~Peres, \emph{Markov Chains and Mixing Times}, 2nd~ed.\hskip 1em plus 0.5em minus 0.4em\relax American Mathematical Society, 2017.

\bibitem{ref24}
J.~Song, C.~Meng, and S.~Ermon, ``Denoising diffusion implicit models,'' in \emph{Proc. Int. Conf. Learn. Represent. (ICLR)}, 2021. [Online]. Available: \url{https://openreview.net/forum?id=St1giarCHLP}

\end{thebibliography}
\end{document}